\documentclass[runningheads]{llncs}

\usepackage{eccv}

\usepackage{eccvabbrv}

\usepackage{graphicx}
\usepackage{booktabs}

\usepackage[accsupp]{axessibility}  

\usepackage{hyperref}

\usepackage{orcidlink}

\usepackage{amsmath,amssymb,amsfonts}
\usepackage{algorithmic}
\usepackage{graphicx}
\usepackage{textcomp}
\usepackage{setspace}
\usepackage{balance}
\usepackage{tabularx}
\usepackage{subcaption}
\usepackage{colortbl}
\usepackage{multirow}
\usepackage{physics}
\usepackage{adjustbox}
\usepackage{threeparttable}
\usepackage{caption}

\usepackage{capt-of}

\usepackage{booktabs}
\usepackage{tikz}
\usetikzlibrary{positioning, fit}

\usepackage{makecell}
\usepackage{float}

\usepackage{xcolor}

\def\BibTeX{{\rm B\kern-.05em{\sc i\kern-.025em b}\kern-.08em
    T\kern-.1667em\lower.7ex\hbox{E}\kern-.125emX}}
\usepackage{annotate-equations}
\definecolor{LightGray}{gray}{0.9}

\begin{document}
\setstretch{0.95}
\title{FlowCon: Out-of-Distribution Detection using Flow-Based Contrastive Learning} 

\titlerunning{FlowCon}

\author{Saandeep Aathreya \orcidlink{0009-0002-2155-1170} \and
Shaun Canavan \orcidlink{0000-0002-1538-476X}
}

\authorrunning{S. Aathreya and S. Canavan}

\institute{University of South Florida\\
\email{\{saandeepaath,scanavan\}@usf.edu}}

\maketitle

\begin{abstract}
  Identifying Out-of-distribution (OOD) data is becoming increasingly critical as the real-world applications of deep learning methods expand. Post-hoc methods modify softmax scores fine-tuned on outlier data or leverage  intermediate feature layers to identify distinctive patterns between In-Distribution (ID) and OOD samples. Other methods focus on employing diverse OOD samples to learn discrepancies between ID and OOD. These techniques, however, are typically dependent on the quality of the outlier samples assumed. Density-based methods explicitly model class-conditioned distributions but this requires long training time or retraining the classifier. To tackle these issues, we introduce \textit{FlowCon}, a new density-based OOD detection technique. Our main innovation lies in efficiently combining the properties of normalizing flow with supervised contrastive learning, ensuring robust representation learning with tractable density estimation. Empirical evaluation shows the enhanced performance of our method across common vision datasets such as CIFAR-10 and CIFAR-100 pretrained on ResNet18 and WideResNet classifiers. We also perform quantitative analysis using likelihood plots and qualitative visualization using UMAP embeddings and demonstrate the robustness of the proposed method under various OOD contexts. Code can be found at \url{https://github.com/saandeepa93/FlowCon\_OOD}.
  
\end{abstract}
\section{Introduction}
\label{sec:intro}



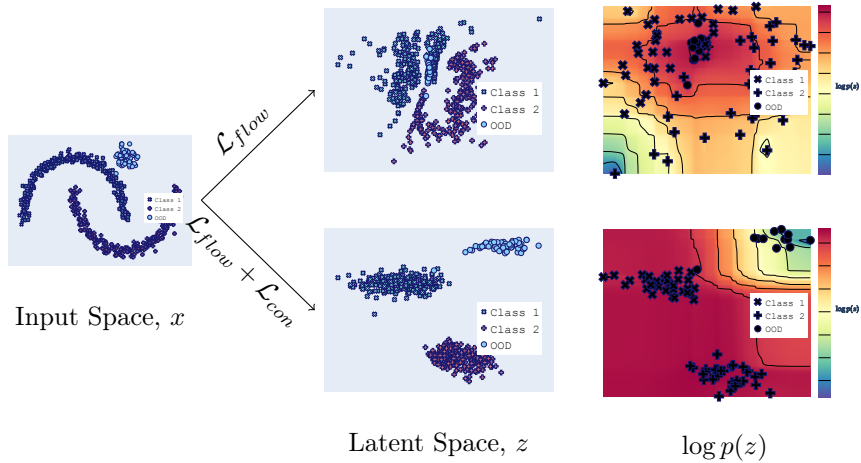
\begin{figure}[t]
  \centering
  \begin{tikzpicture}
    
    \node (centerTopImage) {\includegraphics[width=0.25\textwidth]{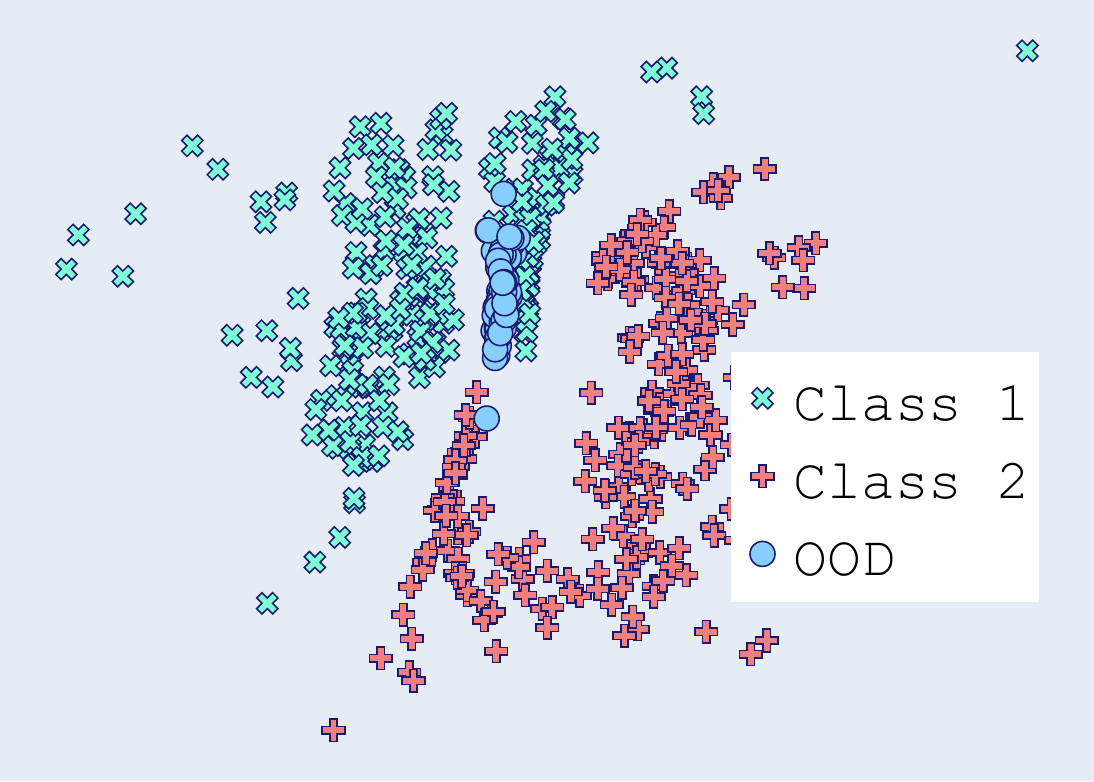}};
    \node[below=0.5cm of centerTopImage] (centerBottomImage) {\includegraphics[width=0.25\textwidth]{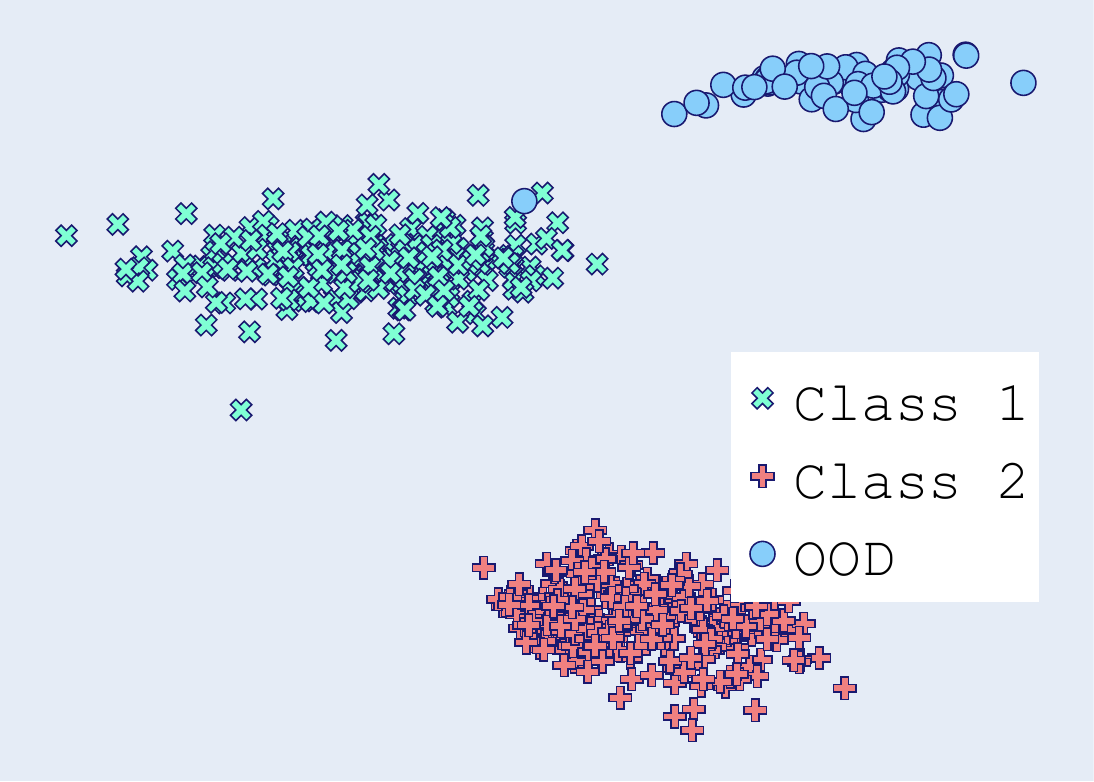}};
    
    \node[fit=(centerTopImage) (centerBottomImage), inner sep=0] (rightImages) {};
    
    \node[left=4.5cm of rightImages.center, anchor=center] (leftImage) {\includegraphics[width=0.2\textwidth]{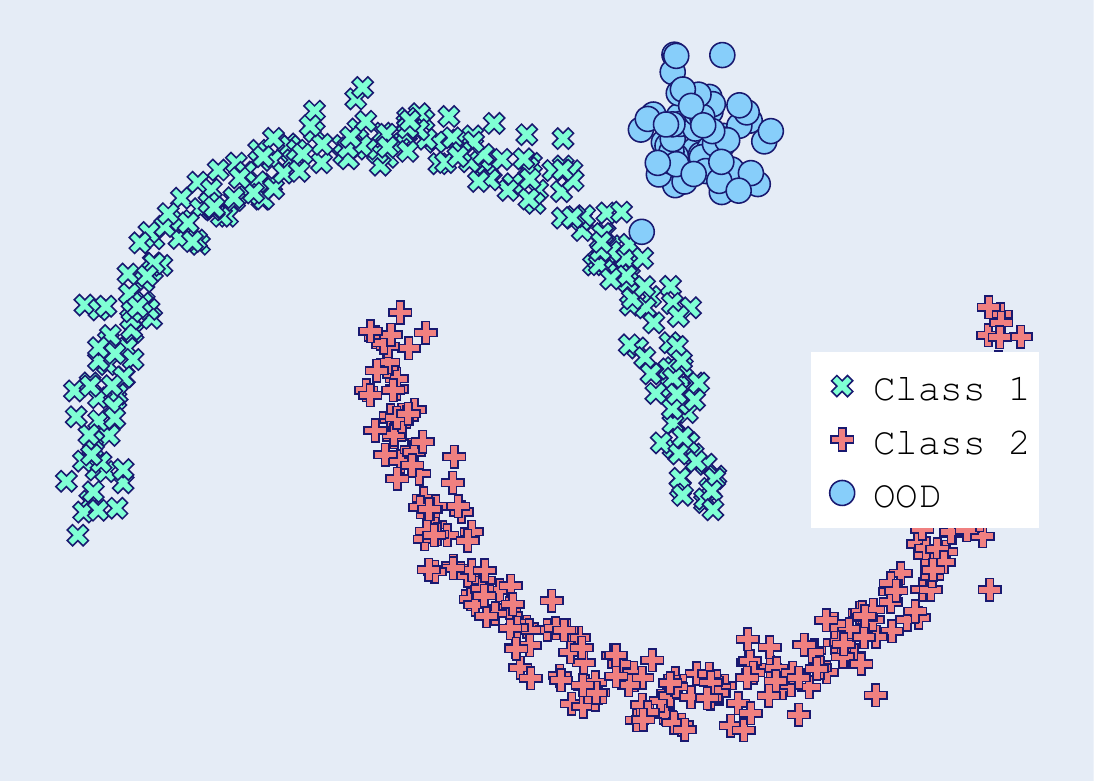}};

    \node[right=0.2cm of centerTopImage] (rightTopImage) {\includegraphics[width=0.3\textwidth]{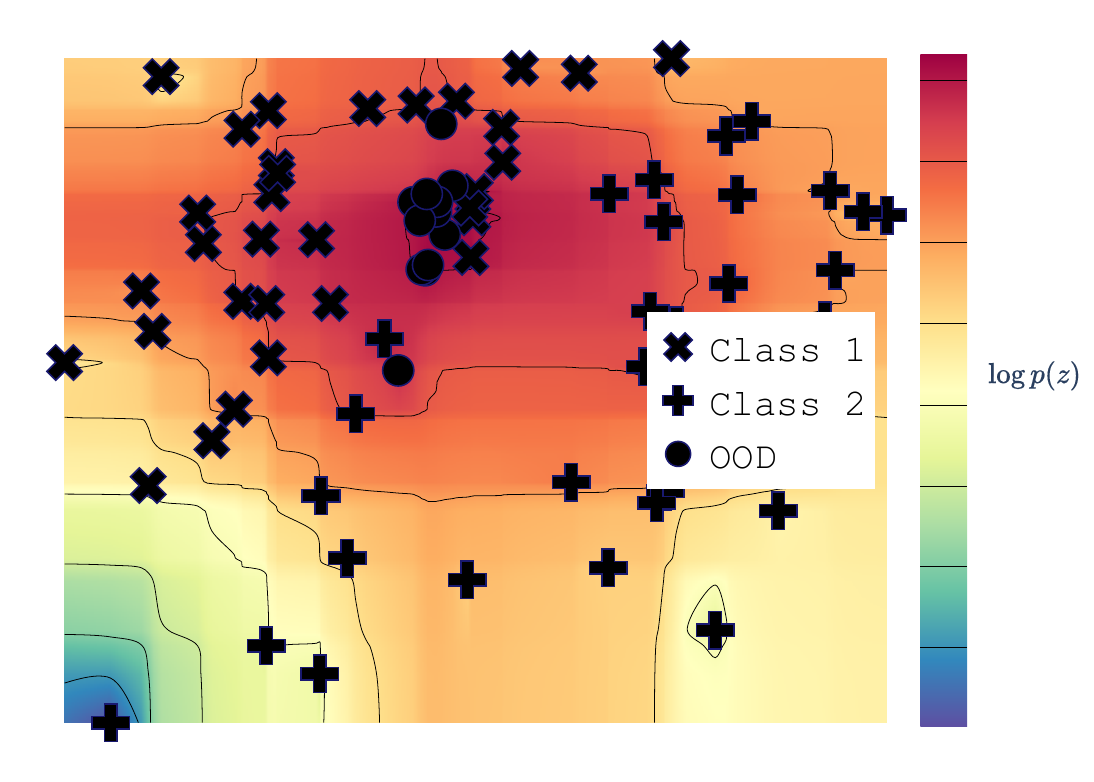}};

    \node[below=0.1cm of rightTopImage] (rightBottomImage) {\includegraphics[width=0.3\textwidth]{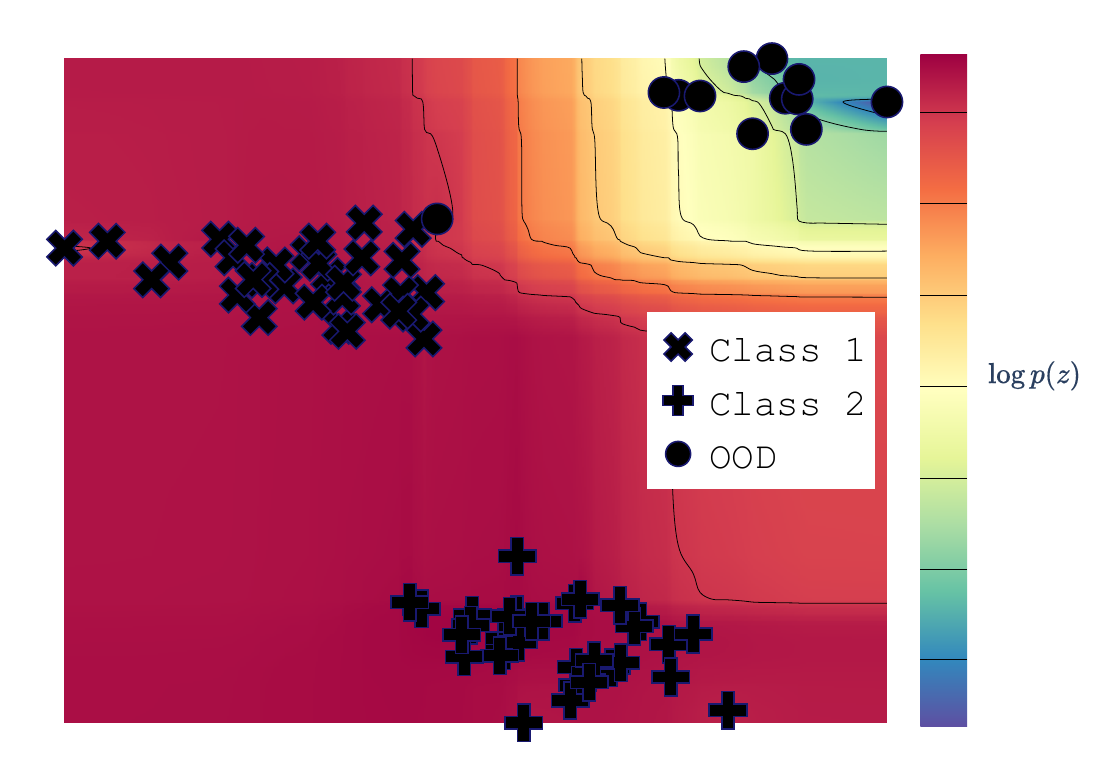}};

    \draw[->] (leftImage.east) -- (centerTopImage.west) node[midway, above, sloped] {$\mathcal{L}_{flow}$};
    \draw[->] (leftImage.east) -- (centerBottomImage.west) node[midway, below, sloped] {$\mathcal{L}_{flow} + \mathcal{L}_{con}$};
    \node[below=0.3cm of leftImage] {Input Space, $x$};
    \node[below=0.3cm of centerBottomImage] {Latent Space, $z$};
    \node[below=0.06cm of rightBottomImage] {$\log p(z)$};
  \end{tikzpicture}


      \caption{\textit{Intuition} behind \textit{FlowCon} on toy moons dataset with OOD samples. Normalizing flows trained without contrastive loss ($\mathcal{L}_{flow}$ only) does not account for the class-specific information in the dataset and transforms the data into a unimodal Gaussian distribution as latent space, $z$. Flow model trained with $\mathcal{L}_{flow} + \mathcal{L}_{con}$ is able to learn class-specific multimodal Gaussian distributions. Consequently, when plotted against $\log p(z)$ using heatmaps, the unimodal Gaussian cluster does not account for OOD samples and assigns high likelihood irrespective of samples it was trained on. Conversely, optimizing $\mathcal{L}_{flow} + \mathcal{L}_{con}$ pushes ID data into the high density region and OOD samples into the low density region.}
      \label{fig:intuition}
    \vspace{-8mm}
\end{figure}

Visual recognition systems are trained under the closed-world assumption that the input distribution at test time remains consistent with the training distribution. This is seldom the case and the model is expected to identify and reject unknown data instances \cite{amodei2016concrete, hendrycks2021unsolved, hendrycks2022x}. In practical scenarios, the test samples may experience gradual distributional shifts and as a result, the model can make arbitrarily incorrect predictions. These shifts can be categorized into \textit{semantic} and \textit{covariate}. Semantic shift (far-OOD) is defined by inclusion of new categories of objects during test time, thereby changing label space. Note that changes in label space naturally impacts the input space as well. On the other hand, covariate shift (near-OOD) is defined by change in the input space only, where the label space remains the same during test time. Collectively, the two present a significant challenge for real world deployment of well-trained systems. This is especially critical in applications such as medical diagnosis \cite{roy2022does, wang2017chestx} and autonomous driving \cite{filos2020can}. 
 
 
Existing methods for OOD detection primarily focus on semantic shift detection. Often fine-tuning the softmax scores of the pretrained classifier through temperature scaling \cite{lee2018simple}, energy scoring \cite{liu2020energy}, or thresholding \cite{sun2021react}. While these methods are simple but powerful, they have been shown to be less effective under near-OOD context \cite{yang2021semantically, winkens2020contrastive}. Other methods leverage large OOD datasets in their training paradigm to make the models sensitive to the OOD test set. Nevertheless, it is unrealistic to make assumption on the vast data space of OOD, which might ultimately introduce bias in the model \cite{shafaei2018less}. 
 
Density-based methods define the score function using the likelihood values, which explicitly model the ID data and identify the low density test data as OOD \cite{lee2018simple, zisselman2020deep, zhang2020hybrid}. Although usually reliable, these methods require preserving the class information by training one model per class \cite{zisselman2020deep}, or retraining the entire classifier under hybrid settings \cite{zhang2020hybrid}. Zisselman \etal \cite{zisselman2020deep} proposed deep residual flows to train one model per class for each layer. This results in significant training requirements given the model and ID dataset being used. For example, deep residual flows trained on CIFAR-100 and pretrained with ResNet-18, will result in 400 flow models ($100\; classes \times 4\;layers$). This is potentially infeasible as the complexity of model and dataset grows. Meanwhile, Zhang \etal \cite{zhang2020hybrid} introduced joint training of the flow model and classifier to represent a multi-modal distribution which can be leveraged for OOD detection. This requires retraining the original classifier models which is not suitable for real-world deployment.

 In this work, we closely follow OOD detection using density estimation. We build upon the principles of generative models, specifically normalizing flows \cite{papamakarios2021normalizing},  to develop a contrastive learning-based approach to tackle the above mentioned constraints. 
 More precisely, in addition to maximizing the log-likelihood of the ID data, we introduce a new loss function which contrastively learns the class specific \textit{distributions} of the ID data. Unlike the conventional contrastive losses \cite{chen2020simple, khosla2020supervised} where the similarity function is typically the cosine similarity of two feature vectors, we leverage \textit{Bhattacharyya coefficient} \cite{bhattacharyya1946measure}, which is designed to measure the similarity between two distributions. The new score emphasizes the network to understand and differentiate between distributions in a contrastive manner. Collectively, the two losses ensure that the network is encouraged to learn semantically meaningful representations enriched with tractable densities. For readability, we name our approach as \textit{FlowCon}. Fig. \ref{fig:intuition} demonstrates the idea of \textit{FlowCon} on a toy dataset. Maximizing the likelihood ($\mathcal{L}_{flow}$) of the dataset without considering the class information results in a latent space with a single Gaussian cluster. Inclusion of the class-preserving contrastive loss ($\mathcal{L}_{flow} + \mathcal{L}_{con}$) pulls the latent Gaussian distribution belonging to same class together, while pushing other distributions away. The figure also shows $\log p(z)$ values using heatmaps to show \textit{FlowCon}'s discriminative properties along with ID/OOD separability.

 To ensure that the original classifier is not modified, we train the flow model and apply the two loss functions ($\mathcal{L}_{flow} + \mathcal{L}_{con}$) on the penultimate layer of the pretrained classifier. As shown by Kirichenko \etal \cite{kirichenko2020normalizing}, training a flow model on deep features focuses on the semantics of the data rather than learning pixel to pixel correlations. To assess the effectiveness of \textit{FlowCon}, we perform quantitative evaluations on benchmark datasets CIFAR-10 and CIFAR-100. We also investigate OOD contexts including far-OOD, near and far-OOD, and near-OOD. The proposed method is competitive or outperforms state-of-the-art OOD detection methods across multiple metrics. 
 To summarize, the contribution of our work is three-fold:
 \vspace{-0.1cm}
 \begin{enumerate}
    \item A new density-based OOD detection technique called \textit{FlowCon} is proposed. We introduce a new loss function $\mathcal{L}_{con}$ which contrastively learns class separability in the probability distribution space. This learning occurs without any external OOD dataset and it operates on fixed classifiers.
    \item The proposed method is evaluated on various metrics - FPR95, AUROC, AUPR-Success, and AUPR-Error and compared against state of the art. We observe that \textit{FlowCon} is competitive or outperforms most methods under different OOD conditions. Additionally, FlowCon is stable even for a large number of classes and shows improvement for high-dimensional features.
    \item Histogram plots are detailed along with unified manifold approximations (UMAP) embeddings \cite{mcinnes2018umap} of the trained FlowCon model to respectively showcase it's OOD detection and class-preserving capabilities. We also show FlowCon's discriminative capabilities. 
\end{enumerate}

\section{Related Work}
\label{sec:related}

\textit{Post-hoc} methods have the benefit of being straightforward to use. They avoid retraining the original classifier or additional training on top of the classifier. Hendrycks \etal \cite{hendrycks2016baseline} proposed early work on OOD detection  by considering the classifier predicted softmax probabilities as the OOD scores. The authors emperically show that the softmax scores of OOD data sufficiently differ from the ID data and therefore forms a baseline for all the subsequent methods. Liang \etal \cite{liang2017enhancing} applied temperature scaling to the softmax probabilities to improve the ID/OOD separability. Additionally, they applied inverse FGSM \cite{goodfellow2014explaining} on the test data that further improved the separability. Liu \etal \cite{liu2020energy} employed a parameter-free softmax caliberation instead of temperature scaling. The authors replace the softmax score with an energy score whose computation forms a theoretical perspective of likelihoods \cite{morteza2022provable}. Sun \etal \cite{sun2021react} designed a truncation technique called ReAct on the penultimate activation layer using a threshold. This truncation threshold was chosen to be the 90th percentile of the ID activations. The authors additionally showcase the compatibility of ReAct with previous techniques such as ODIN \cite{liang2017enhancing}, MSP \cite{hendrycks2016baseline} and Energy \cite{liu2020energy} that further improved the scores. Lee at al. \cite{lee2018simple} use Mahalanobis distance to separate ID/OOD samples. The authors compute class-wise empirical mean and covariance for all the average network activations. This is performed over all the training sets which are then modelled as class conditioned Gaussian distributions. During test time, the score is the maximum weighted Mahalanobis distance between test sample and each distribution. In our experiments, we compare FlowCon with these state-of-the-art techniques and show competitive results across multiple metrics.

\textit{Outlier-based} methods introduce additional training phases extending the pretrained classifier \cite{mohseni2020self, yang2021semantically, lu2023uncertainty, shafaei2018less}. Hendrycks \etal \cite{hendrycks2018deep} proposed exposing the network to outlier samples thereby enhancing its ability to identify and flag test samples that it has not encountered before. Moreover, the authors leverage OOD data to learn heuristics of the ID data without explicitly modelling them. Hornauer \etal \cite{hornauer2023heatmap} propose a heatmap-based approach by attaching a decoder network to a trained classifier layer. Similar to the work from Hendrycks et al., they use outlier OOD samples to define boundaries between ID/OOD samples. A zero-response heatmap output is recognized as ID and a high-response output is categorized as OOD. In our experimental design, we compare FlowCon with the heatmap-based results and follow a similar experimental setup. It is important to note, however, that we did not include them in experiments that require reproducing the results as the OOD dataset they used \cite{torralba200880} has been withdrawn\footnote{https://groups.csail.mit.edu/vision/TinyImages/}.

\textit{Density-based} methods model the ID data without usage of outlier exposure \cite{zong2018deep, abati2019latent, pidhorskyi2018generative, deecke2019image, sabokrou2018adversarially}. Zhang \etal \cite{zhang2020hybrid} propose joint training of classifier and flow models to ensure stronger discriminative and OOD detection capabilities. The authors present a strong motivation to model likelihood of ID data using normalizing flows and present better results on hybrid training as opposed to using fixed classifiers. However, hybrid approaches require retraining the entire classifier which is not suitable for real-world applications. Zisselman \etal \cite{zisselman2020deep} address normality assumptions with Mahalanobis distance \cite{lee2018simple} by training class-wise residual flows \cite{chen2019residual} for each layer of the model. This ensured that the latent features post residual training are a true Gaussian distribution. On the other hand, we train \textit{FlowCon} only on the penultimate layer of the fixed classifier wherein a single model learns class-wise distribution in a supervised manner. We improve upon the ResFlow \cite{zisselman2020deep} model by resolving the training pipeline to consist of only a single model. We compare \textit{FlowCon} with ResFlows and evaluate extensively on OOD detection performance and histogram interpretability (Section \ref{subsec:resflo}). 

\section{Background}
\label{sec:background}

In this section, we briefly introduce the formulation of normalizing flows based on coupling layers and supervised contrastive learning. 
\subsection{Normalizing Flows}
\label{subsec:nf}
Normalizing flows \cite{tabak2013family} are a class of deep generative models \cite{kingma2014semi} that learn to transform a complex distribution, $p_{X}(x)$ to a base distribution, $p_{Z}(z)$ using a sequence of invertible transformations, $z = f(x)$. These transformations are typically in the form of neural networks parametrized by their weights. Using the change of variables formula, the log-likelihood for a datapoint $x$ is maximized by, 
\begin{equation}
    \mathcal{L}_{flow} = -\log p_{X}(x) = -\bigg[\log p_{Z}(f(x)) + \log \bigg | \det \frac{\partial f(x)}{\partial x} \bigg | \bigg].
\label{eqn:cv}
\end{equation}
The base distribution $p_{Z}(z)$ is commonly chosen to be standard Gaussian. 
To satisfy the properties required for Equation \ref{eqn:cv}, $f$ has additional constraints in model architecture. More specifically, $f$ should be bijective and the Jacobian determinant of $f$ should be easy to compute. Addtionally, due to bijective properties of flow models, $x \in \mathbb{R}^d$, and $z \in \mathbb{R}^d$ have same dimensions with $z \sim \mathcal{N}(0, I)$. We refer the readers to the works of Papamakarios et. al. \cite{papamakarios2021normalizing} and Dinh et. al. \cite{dinh2016density} for a comprehensive introduction.

\subsection{Supervised Contrastive Learning (SCL)}
\label{subsec:scl}
SCL \cite{khosla2020supervised} is a family of representation learning frameworks that aim at learning the most informative deep embeddings of images. Given a set of $I$ data instances $\{x_i, y_i\}_{i=1,...,I}$  in a multi-viewed batch, SCL takes the following form 
\begin{equation}
\small
    \mathcal{L}_{supcon} = \sum_{i \in I} \frac{-1}{|P(i)|}\sum_{p \in P(i)} \log \frac{S(z_i, z_p)/\tau}{\sum_{a \in A(i)}S(z_i, z_a)/\tau)}.
    \label{eqn:scl}
\end{equation}
Here, $S(z_i, z_j) = \exp(z_i \cdot z_p)$
is the similarity function, $z_i = f(x_i)$ is the latent embedding of the anchor image $x_i$, $P$ is the set of all positives where $y_i = y_p$, except $z_i$, and $A(i) \equiv I \symbol{92} i$ is the set of all positives and negatives, except $z_i$. SCL has shown strong results in creating successful semantic representations of input datasets \cite{wen2022self}. Moreover, Winkens \etal \cite{winkens2020contrastive} demonstrated that the contrastive approach to classification further improves OOD detection capabilities of the classifier. In general, Equation \ref{eqn:scl} aims to minimize distances between data pairs of similar classes while maximizing the distance between dissimilar classes using the dot product between the feature vectors. For a more detailed introduction, we refer the reader to works from Khosla et. al. \cite{khosla2020supervised} and Frosst et. al. \cite{frosst2019analyzing}.
\section{Flow-based Contrastive Learning}
\label{sec:method}

\begin{figure}[t]
    \centering
    \includegraphics[width=0.85\linewidth, height=4.3cm]{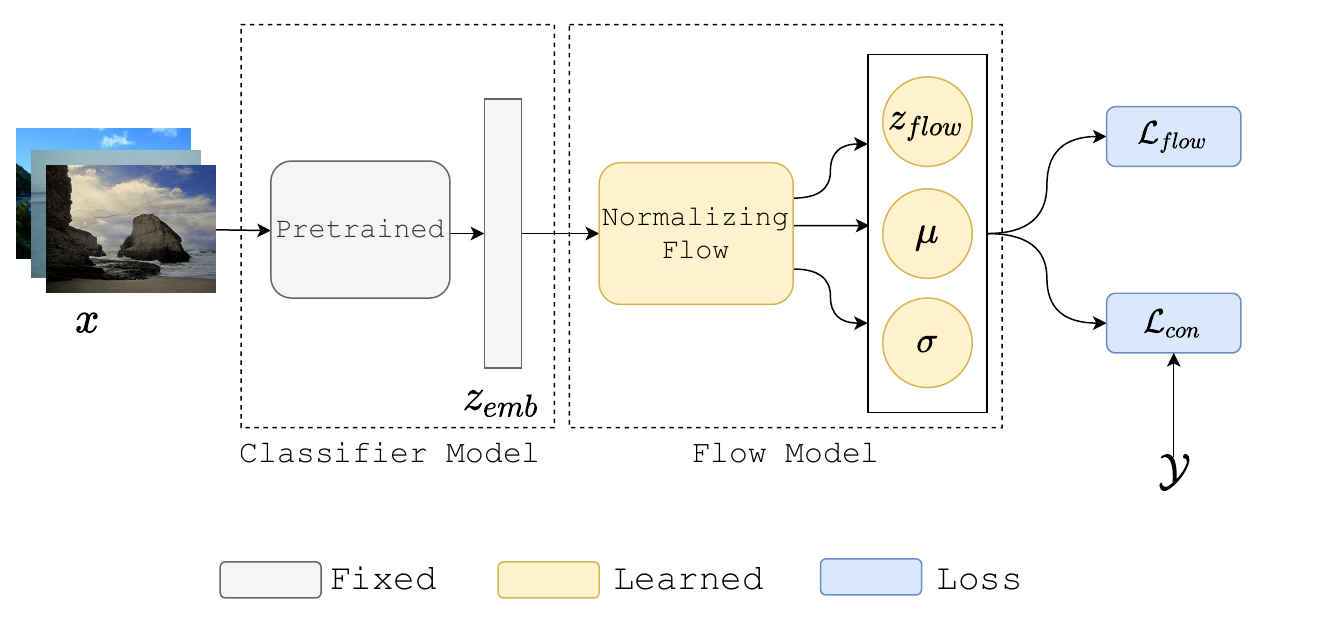}
    \caption{Training pipeline of \textit{FlowCon}. Given an input image, $x$, the pretrained classifier first extracts deep features, $z_{emb}$. The flow model then operates on $z_{emb}$ to obtain the latent vector $z_{flow}$, and its corresponding distribution, $\mathcal{N}(\mu, \sigma)$. The loss $\mathcal{L}_{flow}$ maximizes the likelihood of $z_{flow}$ on $\mathcal{N}(\mu, \sigma)$, and simultaneously, $\mathcal{L}_{con}$  ensures high inter-class separability and low intra-class separability among the distribution $\mathcal{N}(\mu, \sigma)$ in a contrastive fashion. }
    \label{fig:arch}

\end{figure}



Given an input image $x \in \mathbb{R}^D$ in the batch, \textit{FlowCon} initially extracts high dimensional deep features using a pretrained classifier network. This new embedding, $z_{emb} \in \mathbb{R}^d$ is now used as input to the normalizing flow model to obtain the latent embedding  $z_{flow} \in \mathbb{R}^d, \mu \in \mathbb{R}^d$, and $\sigma \in \mathbb{R}^d$. 
Details regarding training and classification are given in Sections \ref{subsec:training} and \ref{subsec:inference}, respectively. See Fig. \ref{fig:arch} for an overview of the proposed architecture.


\subsection{Training}
\label{subsec:training}
To strike a good balance between discriminative and semantic properties of latent embeddings $z_{flow}$, we propose to effectively combine Equations \ref{eqn:cv} and \ref{eqn:scl}. Instead of a naive merging of equations, we use an efficient similarity measure, $S_{flow}$, that uses the likelihood information, $p_Z(z)$, obtained in Equation \ref{eqn:cv}. This reduces the high-dimensional vector dot-product to a simple scalar product of likelihoods. The new similarity function $S_{flow}$ for a given batch is then written as

\begin{equation}
    S_{flow}(z_i, z_j, \mathcal{N}_i) = \exp\bigg((p_Z(z_i|\mathcal{N}_i) \cdot p_Z(z_j|\mathcal{N}_i))^{\tau_1}\bigg)
    \label{eqn:sim_new}
\end{equation}
where, 
\begin{equation}
    p_Z(z_i| \mathcal{N}_i) = \frac{1}{\sigma\sqrt{2 \pi}} \exp \bigg[{\frac{-1}{2}\bigg(\frac{z_i - \mu_i)}{\sigma_i}\bigg)^2}\bigg],
\end{equation}
and $\tau_1$ is a hyperparameter. Here, $p_Z(z_i | \mathcal{N}_i)$ denotes the likelihood of latent embedding $z_i$ belonging to distribution $\mathcal{N}_i$. Note that $p_Z(z_i | \mathcal{N}_i)$ is obtained from Equation \ref{eqn:cv} (as $\log p_Z(z)$). For ease of sampling in normalizing flows, $\mathcal{N}_i$ resolves to unit hypersphere. However, since our aim is to learn class-specific distributions, we let the flow network learn the distributions (see Fig \ref{fig:arch}). \footnote{The idea of learned $\mu$ and $\sigma$ was first adopted in the GLOW implementation in https://github.com/openai/glow}  

Here, $p_Z(z_i|\mathcal{N}_i) \cdot p_Z(z_j|\mathcal{N}_i)$ should yield a high value when $y_i = y_j$. Conversely, the product should yield a lower value when $y_i \neq y_j$. This is analogous to traditional SCL,  where, the higher the dot product between the latent vectors, the more similar the images are, and therefore, the closer they are in the feature space. Conversely, if the dot product is low, the images are dissimilar, and hence, farther apart in the feature space. Therefore, contrastive loss fits naturally in this context.  
Additionally, the term inside $\exp$ in Equation \ref{eqn:sim_new} is the generalized form of Bhattacharyya coefficient \cite{bhattacharyya1946measure} when the hyperparameter $\tau_1=0.5$.
Finally, combining Equations \ref{eqn:scl} and \ref{eqn:sim_new} together we get, 
\begin{equation}
    \mathcal{L}_{con} = \sum_{i \in I} \frac{-1}{|P(i)|}\sum_{p \in P(i)} \log \frac{S_{flow}(z_i, z_p, \mathcal{N}_i)/\tau_2}{\sum_{a \in A(i)}S_{flow}(z_i, z_a, \mathcal{N}_i)/\tau_2}.
    \label{eqn:flowcon}
\end{equation}
It is important to note that unlike the traditional contrastive learning methods, besides the latent vector $z_i$, the distribution $\mathcal{N}_i$ also serves as anchor. 

Overall, optimizing $\mathcal{L}_{con}$ has an intuitive interpretation as it learns distributions $\mathcal{N}$ in a contrastive manner. On the other hand, $\mathcal{L}_{flow}$ learns embeddings $z_{flow}$ that belongs to the distributions learned by optimizing $\mathcal{L}_{con}$.
The loss function in Equation \ref{eqn:cv} remains the same and we optimize Equations \ref{eqn:cv} and \ref{eqn:flowcon} concurrently with a scaling constant $\lambda$ as $\mathcal{L} = \mathcal{L}_{con} + \lambda \mathcal{L}_{flow}.$
\subsection{OOD Detection with \textit{FlowCon}}
\label{subsec:inference}
Ideally, at the end of training, we obtain a distribution, $\mathcal{N}$, for each data point in the training set $\mathcal{X} = \{x_1, x_2, ..., x_n\}$. For $n$ data points, we will obtain $\mathcal{N}_\mathcal{X}=\{\mathcal{N}_1, \mathcal{N}_2, ..., \mathcal{N}_n\}$ distributions. To enable downstream tasks, such as OOD detection, we simplify the task of dealing with $n$ distributions by reducing them to a smaller set of $k$ distributions, where $k$ is the number of classes.
We perform this by taking the emperical mean of the distributions per class. Therefore, the parameters $\mu_c$ and $\sigma_c$ of the  distribution $\mathcal{N}_c$ for a class $c$ is computed as 
\begin{equation}
    \mu_c = \frac{1}{|\mathcal{X}_c|}\sum_{i \in \mathcal{X}_c}\mu_i;   
    \sigma_c = \frac{1}{|\mathcal{X}_c|}\sum_{i \in \mathcal{X}_c}\sigma_i,
\end{equation}
where $\mathcal{X}_c \equiv \{i \in \mathcal{X} : y_i = c\}$ is the total instances in the training set with class label $c$. Repeating the process for each class, we get a total of $k$ distributions for the training set $\mathcal{X}$, given by $\mathcal{N}_\mathcal{X}=\{\mathcal{N}_1, ..., \mathcal{N}_k\}$. To compute the score, $S$ for OOD detection on a test sample $z_{test}$, we simply compute its likelihood on all distributions and take the maximum value $S(x_{test}) = \max_{i\in \{1, ..., k\}}p_Z(z_{test}|\mathcal{N}_{y=i}).$

\section{Experiments}
\label{sec:exp}

\subsection{Setup}
 
 \subsubsection{Dataset and Models.}
 \label{subsubsec:dataset}
 We use CIFAR-10 \cite{krizhevsky2009learning} and CIFAR-100 \cite{krizhevsky2009learning} datasets as in-distribution ($D_{in})$. For OOD datasets ($D_{ood}$), we rely on 6 external test sets: iSUN \cite{xu2015turkergaze}, LSUN-Crop \cite{yu2015lsun}, LSUN-Resize \cite{yu2015lsun}, SVHN \cite{netzer2011reading}, Textures \cite{cimpoi2014describing}, and Places365 \cite{zhou2017places}. For the pretrained classifier, we use ResNet18 \cite{he2016deep} and WideResNet \cite{zagoruyko2016wide} with depth 40 and width 2 which have been trained on both CIFAR-10 and CIFAR-100. This allows us to evaluate FlowCon on various scales of penultimate feature dimensions ($512$ for ResNet18 and $128$ for WideResNet). 

\subsubsection{Evaluation Metrics.}
To provide the best metrics independent of a particular OOD score threshold, we evaluate \textit{FlowCon} using four standard metrics, AUROC, AUPR-Success (AUPR-S), AUPR-Error (AUPR-E), and FPR at 95\% TPR (FPR-95). AUROC integrates under the reciever operating characteristics (ROC) curve measuring the model's performance across various threshold settings. AUPR-S is area under the precision-recall curve. It focuses on performance of the model in correctly classifying ID samples. Conversely, AUPR-E is the perfomance of the model in correctly identifying OOD samples. FPR-95 is a measure of how often the model incorrectly identifies an OOD sample as ID when it is correctly identifying 95\% of ID samples. Similar to prior works \cite{hendrycks2018deep, liu2020energy}, the entire test set of ID samples are considered and the number of OOD samples are randomly selected to be one-fifth of ID test set. The results are averaged for all OOD datasets.

\subsubsection{Implementation Details}
We follow the experimental setup of Hornauer et al. \cite{hornauer2023heatmap}, wherein we build \textit{FlowCon} on top of the pretrained Resnet18 and WideResNet.\footnote{Classifier weights obtained from https://github.com/jhornauer/heatmap\_ood} For the flow model, we adopt a  standard RealNVP architecture with $8$ coupling blocks and a single flow layer. \footnote{https://github.com/PolinaKirichenko/flows\_ood} For ResNet18, we train \textit{FlowCon} on the fixed $512$ dimensional penultimate features and $128$ dimensional features for WideResNet. The multitask loss function $\mathcal{L}_{total}$ is optimized using Adam optimizer \cite{kingma2014adam} with a fixed learning rate of $1e-5$ and weight decay of $1e-5$. For all experiments, the flow model is trained for $700$ epochs with a batch size of $64$ with an image size of $32 \times 32$. Moreover, we emperically find $\lambda$ value to be $0.07$. We fix the $\mathcal{L}_{con}$ hyperparameters $\tau_1$ and $\tau_2$ at $1.5$ and $0.1$, respectively.


\subsubsection{\textit{FlowCon} vs. Competitive baselines}
We compare \textit{FlowCon} with methods that operate only on fixed classifiers. These include: 
\begin{itemize}
    \item \textit{Post-hoc} methods that either calibrate or scale the softmax scores like MSP \cite{hendrycks2016baseline}, ODIN \cite{liang2017enhancing}, Mahalanobis \cite{lee2018simple}, Energy \cite{liu2020energy}, and ReAct \cite{sun2021react}. Note that ODIN and Mahalanobis require finetuning the hyperparameters based on external OOD datasets. For all the post-hoc methods, we follow the hyperparameter selection which is consistent with Hornauer et al. \cite{hornauer2023heatmap}.
    
    
    \item \textit{Outlier-based}. Since our approach performs additional training, we also consider methods that train on outlier OOD datasets and a flow-based method. For outlier trained methods, we compare with the heatmap-based approach as proposed by Hornauer \etal \cite{hornauer2023heatmap} which showed state-of-the-art results.

    \item \textit{Flows}. We consider \textit{residual flows} \cite{zisselman2020deep} and train the classifier features for all layers in a class-wise manner. Similar to Mahalanobis, residual flows additionally trains a regressor fine-tuned on OOD test sets to predict the scores. In contrast, \textit{FlowCon} operates on penultimate features without a post inference regressor. We show that \textit{FlowCon} outperforms residual flows across the majority of experiments, especially under challenging OOD contexts. 
\end{itemize}



\subsection{Result on  OOD Contexts}
\label{subsec:res}
We study the OOD detection capabilities of \textit{FlowCon} under three type of ($D_{in}$, $D_{ood}$) pairs. These pairs cover a broad spectrum of OOD test instances encountered in the real-world. Similar to the works of Winkens et al. \cite{winkens2020contrastive}, each ID dataset will be paired against:
\vspace{-0.5cm}
\subsubsection{\textit{Far-OOD}.} We compare both CIFAR-10 and CIFAR-100 against the six external datasets ($D_{ood}$) listed in Section \ref{subsubsec:dataset}. These far-OOD experiment pairs are characterized by semantic shifts.
\begin{table}[t]
    \setlength\tabcolsep{1pt}
    \begin{center}
    \caption{\textit{Far-OOD:} Comparison of OOD detection performance during only semantic shift. $\ast$ Uses OOD data to finetune the hyperparameters. ${\dagger}$ Uses OOD dataset for training. ${\ddag}$ Explicitly uses flow models. The results are averaged over the number of OOD test sets ($D_{ood}$) mentioned in Section \ref{sec:exp}.}
    \label{tab:far_ood_results}
    \vskip -1em
    \begin{adjustbox}{width=0.9\textwidth, totalheight=0.6\textheight, keepaspectratio}
    \begin{tabular}{clcccc}
    \toprule
        $D_{in}$ & \multirow{2}{*}{Method} & \multirow{2}{*}{AUROC $\uparrow$} & \multirow{2}{*}{AUPR-S $\uparrow$} & \multirow{2}{*}{AUPR-E $\uparrow$} & \multirow{2}{*}{FPR-95 $\downarrow$} \\
        (model) & & & & & \\
        \midrule
        \multirow{7}{*}{\makecell{CIFAR-10 \\ (ResNet)}} & MSP \cite{hendrycks2016baseline} & 90.72 & 97.89 & 63.48 & 55.21 \\
        & ODIN* \cite{liang2017enhancing}   & 88.33 & 96.67 & 71.49 & 38.35 \\
        & Mahalanobis* \cite{lee2018simple}  & 92.33 & 98.29 & 71.30 & 39.52 \\
        & Energy \cite{liu2020energy} & 91.72 & 97.90 & 72.12 & 37.97 \\
        & ReAct  \cite{sun2021react} & 91.71 & 97.89 & 72.55 & 36.52 \\
        & ResFlow$^{\ddag}$ \cite{zisselman2020deep} & 95.6 & \underline{99.35} & 82.82 & \textbf{13.22} \\
        & Heatmap$^{\dagger}$ \cite{hornauer2023heatmap} & \underline{96.47} & 99.17 & \underline{83.73} & \underline{15.37} \\
        & FlowCon (Ours) & \textbf{97.19} & \textbf{99.43} & \textbf{85.65} & 16.26 \\
        
        \midrule
        \multirow{8}{*}{\makecell{CIFAR-10 \\ (WideResNet)}} & MSP \cite{hendrycks2016baseline} & 91.48 & 98.18 & 63.47 & 56.77 \\
        & ODIN* \cite{liang2017enhancing}  & 95.01 & 98.68 & 84.39 & 21.09 \\
        & Mahalanobis* \cite{lee2018simple}  & 92.03 & 98.09 & 75.44 & 32.73 \\
        & Energy \cite{liu2020energy} & 94.91 & 98.75 & 80.89 & 24.26 \\
        & ReAct \cite{sun2021react} & 51.92 & 85.46 & 17.53 & 97.12 \\
        & ResFlow$^{\ddag}$ \cite{zisselman2020deep} & 81.58 & 66.09 & \underline{86.78} & 49.11 \\
        & Heatmap$^{\dagger}$ \cite{hornauer2023heatmap} & \underline{96.36} & \textbf{99.07} & 86.73 & \textbf{14.06} \\
        & FlowCon (Ours) & \textbf{96.42} & \underline{98.84} & \textbf{86.90} & \underline{19.10} \\
        
        \midrule
        \multirow{7}{*}{\makecell{CIFAR-100 \\ (ResNet)}} & MSP \cite{hendrycks2016baseline} & 79.29 & 95.04 & 40.34 & 76.58 \\
        & ODIN* \cite{liang2017enhancing} & 83.28 & 95.96 & 48.74 & 67.96 \\
        & Mahalanobis* \cite{lee2018simple}  & 73.46 & 93.00 & 35.90 & 79.46 \\
        & Energy \cite{liu2020energy}  & 82.07 & 95.71 & 43.92 & 74.45 \\
        & ReAct \cite{sun2021react} & 84.22 & 96.27 & 49.08 & 67.78 \\
         & ResFlow$^{\ddag}$ \cite{zisselman2020deep} & 85.12 & 71.45 & \textbf{67.89} & \underline{42.55} \\
        & Heatmap$^{\dagger}$ \cite{hornauer2023heatmap} & \underline{86.74} & \underline{96.49} & \underline{58.78} & 52.73 \\
        & FlowCon (Ours) & \textbf{88.22} & \textbf{96.85} & \textbf{67.89} & \textbf{41.85} \\
        
        \midrule
        
        \multirow{8}{*}{\makecell{CIFAR-100 \\ (WideResNet)}} & MSP  \cite{hendrycks2016baseline} & 65.31 & 90.38 & 26.21 & 88.45 \\
        & ODIN* \cite{liang2017enhancing} & 79.43 & 94.60 & 43.98 & 73.19 \\
        & Mahalanobis* \cite{lee2018simple}  & 73.99 & 92.58 & 43.80 & 68.45 \\
        & Energy  \cite{liu2020energy}  & 77.11 & 93.95 & 39.07 & 78.03 \\
        & ReAct \cite{sun2021react} & 80.74 & 95.24 & 48.04 & 67.47 \\
        & ResFlow$^{\ddag}$ \cite{zisselman2020deep} & \textbf{88.58} & 62.36 & \textbf{89.17} & 65.77 \\
        & Heatmap$^{\dagger}$ \cite{hornauer2023heatmap}  & \underline{85.98} & \underline{95.96} & \underline{61.14} & \textbf{49.86} \\  
        & FlowCon (Ours) & 83.62 & \textbf{96.60} & 53.34 & \underline{60.28} \\

    \bottomrule
    \end{tabular}
    \end{adjustbox}
    \end{center}
    \vspace{-5mm}

\end{table}
The performance of \textit{FlowCon} along with its benchmarks is listed in Table \ref{tab:far_ood_results}. We observe that \textit{FlowCon} performs exceedingly well for CIFAR-10 and CIFAR-100 pretrained on ResNet18 model. This indicates that \textit{FlowCon} is robust even for a higher number of classes. For WideResNet model trained on CIFAR-10, \textit{FlowCon} reports the highest performance for AUROC and AUPR-E with $96.2$ and $86.90$, respectively. It achieves second-best performance for AUPR-S and FPR-95 reported as $98.84$ and $19.10$, respectively, after the Heatmap approach \cite{hornauer2023heatmap}. For WideResNet on CIFAR-100, \textit{FlowCon} obtains the best performance on AUPR-S with a score of $96.60$ while retaining competitive measures across all other metrics. For instance, ResFlow secures the best AUROC and AUPR-E, however, its AUPR-S remains low, indicating more misclassified ID samples in an effort to filter out OOD samples. Heatmap uses external OOD datasets during training and remains consistent in it's performance with the lowest FPR-95. We provide results on individual OOD datasets in the supplementary material.

\begin{table}[t]
    \begin{center}
    \caption{\textit{Near-far and near-OOD}. Comparison of OOD detection performance during both semantic and covariate shift. $\ast$ Uses OOD data to finetune the hyperparameters. ${\dagger}$ Uses OOD dataset for training. ${\ddag}$ Explicitly uses flow models.}
    \label{tab:mixed_ood_results}
    \small{
    \begin{adjustbox}{width=0.8\textwidth, totalheight=0.5\textheight, keepaspectratio}
    
    \begin{tabular}{cclcccc}
    \toprule
        $D_{in}$ & \multirow{2}{*}{$D_{ood}$} & \multirow{2}{*}{Method} & \multirow{2}{*}{AUROC $\uparrow$} & \multirow{2}{*}{AUPR-S $\uparrow$} & \multirow{2}{*}{AUPR-E $\uparrow$} & \multirow{2}{*}{FPR-95 $\downarrow$} \\
        (model) & & & & & & \\
        \midrule
        \multirow{6}{*}{\makecell{CIFAR-10 \\ (ResNet)}} & \multirow{6}{*}{CIFAR-100} & MSP \cite{hendrycks2016baseline} & \underline{86.45} & \underline{96.49} & 53.15 & 65.95 \\
        & & ODIN* \cite{liang2017enhancing} & 64.79 & 89.86 & 24.32 & 90.85 \\
        & & Mahalanobis* \cite{lee2018simple} & 63.90 & 88.83 & 29.57 & 82.55 \\
        & & Energy \cite{liu2020energy} & 85.60& 95.87& 57.66& 55.2\\
        & & ReAct \cite{sun2021react} & 85.36  & 95.76 & 57.51 & \underline{54.85}\\
        & & ResFlow$^{\ddag}$  \cite{zisselman2020deep} & 76.40 & 26.23 & \underline{66.23} & 67.2\\
        & & FlowCon (Ours) & \textbf{93.97} & \textbf{98.74} & \textbf{73.84} & \textbf{35.95}\\
        
        \midrule
        \multirow{6}{*}{\makecell{CIFAR-10 \\ (WideResNet)}} &  \multirow{6}{*}{CIFAR-100} & MSP \cite{hendrycks2016baseline} & \underline{86.47} & \underline{96.87}& 52.43& 67.65\\
        & & ODIN* \cite{liang2017enhancing} & 71.89& 91.48& 33.74& 80.6\\
        & & Mahalanobis* \cite{lee2018simple} & 65.40& 88.95& 29.83& 83.4\\
        & & Energy \cite{liu2020energy} & \textbf{87.50} & 96.84 & \underline{60.90} & \textbf{52.85} \\
        & & ReAct \cite{sun2021react} & 63.7 & 90.36 & 22.31 & 92.25 \\
        & & ResFlow$^{\ddag}$ \cite{zisselman2020deep} & 53.38 & 12.01 & \textbf{90.59} & 94.53 \\
        & & FlowCon (Ours) & 85.24 & \textbf{96.90} & 57.77 & \underline{56.9}\\

    \bottomrule
    \end{tabular}
    \end{adjustbox}
    }
    \end{center}
    \vspace{-5mm}

\end{table}
\begin{table}
    \begin{center}
    \caption{\textit{Near-OOD} Comparison of OOD detection performance during only covariate shift. $\ast$ Uses OOD data to finetune the hyperparameters. ${\dagger}$ Uses OOD dataset for training. ${\ddag}$ Explicitly uses flow models.}
    \label{tab:near_ood_results}
    \small{
    \begin{adjustbox}{width=0.8\textwidth, totalheight=0.5\textheight, keepaspectratio}
    
    \begin{tabular}{cclcccc}
    \toprule
        $D_{in}$ & \multirow{2}{*}{$D_{ood}$} & \multirow{2}{*}{Method} & \multirow{2}{*}{AUROC $\uparrow$} & \multirow{2}{*}{AUPR-S $\uparrow$} & \multirow{2}{*}{AUPR-E $\uparrow$} & \multirow{2}{*}{FPR-95 $\downarrow$} \\
        (model) & & & & & & \\
        \midrule
         \multirow{6}{*}{\makecell{CIFAR-100 \\ (ResNet)}} &  \multirow{6}{*}{CIFAR-10} & MSP \cite{hendrycks2016baseline}  & 76.53& 94.25& 35.28& 82.5\\
        & & ODIN* \cite{liang2017enhancing} & 60.46& 88.79& 20.91& 93.01\\
        & & Mahalanobis*  \cite{lee2018simple} & 42.54& 81.21& 13.53& 98.6\\
        & & Energy \cite{liu2020energy} & \underline{77.06} & \underline{94.26}& 36.00& 81.15\\
        & & ReAct \cite{sun2021react} & 50.49 & 73.63 & 16.7 & 95.2 \\
        & & ResFlow$^{\ddag}$ \cite{zisselman2020deep} & 58.29 & 46.34 & \underline{47.48} & \underline{79.0} \\
        & & FlowCon (Ours) & \textbf{82.80}& \textbf{95.79}& \textbf{48.79}& \textbf{67.6}\\
        
        \midrule
        \multirow{6}{*}{\makecell{CIFAR-100 \\ (WideResNet)}} & \multirow{6}{*}{CIFAR-10} & MSP \cite{hendrycks2016baseline}   & \underline{72.85} & \textbf{93.46}& 31.54& 85.75\\
        & & ODIN* \cite{liang2017enhancing} & 62.00& \underline{89.39} & 22.08& 92.05\\
        & & Mahalanobis*  \cite{lee2018simple} & 42.97& 81.13& 13.7& 98.35\\
        & & Energy \cite{liu2020energy} & \textbf{74.30} & \textbf{93.46} & \underline{32.94} & \underline{83.6}\\
        & & ReAct \cite{sun2021react} & 49.08 & 82.9 & 16.33 & 95.01 \\
        & & ResFlow$^{\ddag}$ \cite{zisselman2020deep} & 59.22 & 18.08 & \textbf{92.34} & 90.81 \\  

        & & FlowCon (Ours) & 67.03& 90.16& 27.86& \textbf{82.85}\\

    \bottomrule
    \end{tabular}
    \end{adjustbox}
    }
    \end{center}
    \vspace{-10mm}
\end{table}
\subsubsection{ \textit{Mixed near-} \& \textit{far-OOD}.} CIFAR-10 ($D_{in}$) is assessed against CIFAR-100 ($D_{ood}$), which is regarded as a mixed near- and far-OOD scenario due to the shared classes between the two datasets. This particular pairing involves both semantic and covariate shifts within the test data. Table \ref{tab:mixed_ood_results} compares our approach with post-hoc methods and ResFlow. Note that Heatmap is evaluated only on far-OOD context in this work since the dataset primarily used by outlier training methods is 80 million TinyImages \cite{torralba200880}, which has been withdrawn from further usage\footnote{https://groups.csail.mit.edu/vision/TinyImages/}. \textit{FlowCon} demonstrates the best performance for ResNet18 model across all metrics. For WideResNet, it acheives the best AUPR-S ($96.90$) and second best FPR-95 ($56.90$) scores. Energy-based thresholding \cite{liu2020energy} achieves the highest AUROC and AUPR-E. Once again, ResFlow reports the lowest FPR-95 which is obtained at the cost of poor AUPR-S measure.

\subsubsection{ \textit{Near-OOD.}}
\label{subsubsec:near_ood}
When CIFAR-100 ($D_{in}$) is tested against CIFAR-10 ($D_{ood}$), it is treated as a near-OOD context because the test set experiences covariate shift without any semantic alterations. Some literature treat covariate shift in test set as ID data \cite{yang2021semantically}. We believe that correctly classifying a test data under extreme covariate shift reflects a classifier's generalization ability, thus its identification is crucial. As can be seen in Table \ref{tab:near_ood_results}, \textit{FlowCon} exhibits superior performance over other methods by attaining the highest score on ResNet18 model and reports the best outcome on FPR-95 ($82.85$) for WideResNet model. Overall, the scores for WideResNet model are shared by MSP, ODIN, Energy, ResFlow, and FlowCon, further highlighting the challenges with near-OOD scenarios. Interestingly, for WideResNet, we note that post-hoc methods in general display moderately better performance for near-OOD scenarios as opposed to far-OOD and mixed-OOD where training based methods (ResFlow/Heatmap) displayed optimum performance. Moreover, the difference in performance between ResNet and WideResNet for \textit{FlowCon} is, in part, due to the feature dimensions on which it operates on. Coupling-based flow architectures like RealNVP \cite{dinh2016density} and Glow \cite{kingma2018glow} have shown promising results on higher dimensional data as opposed to low-dimensional features \cite{reyes2023testing} (e.g., $128$ for WideResNet). 


\subsection{Likelihood Plots}
\begin{figure}
    \centering
    \begin{minipage}{.45\textwidth}
        \centering
        \begin{subfigure}[t]{0.45\textwidth}
            \centering
            \includegraphics[width=\textwidth, height=2.5cm]{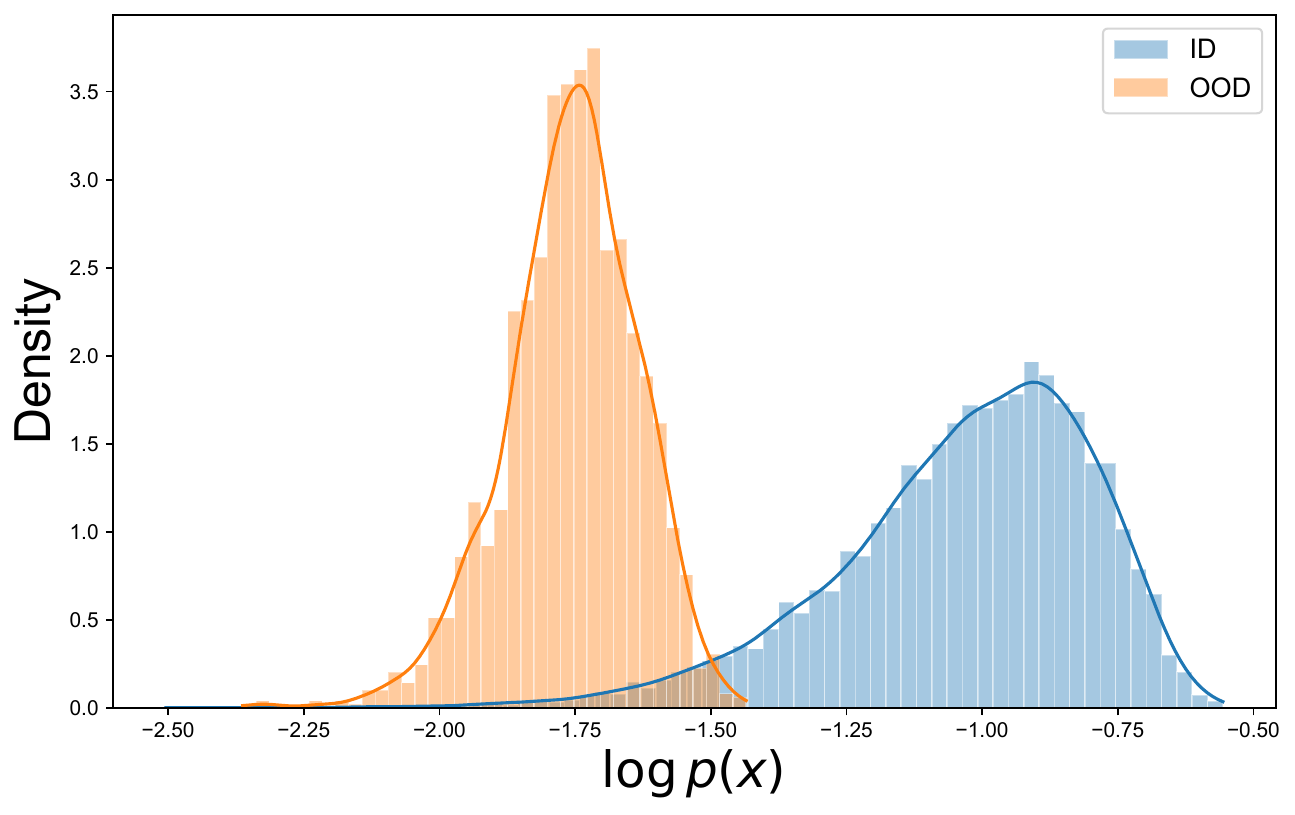}
            \caption*{}
            \label{fig:cifar10_3_ll}
        \end{subfigure}%
        \begin{subfigure}[t]{0.45\textwidth}
            \centering
            \includegraphics[width=\textwidth, height=2.5cm]{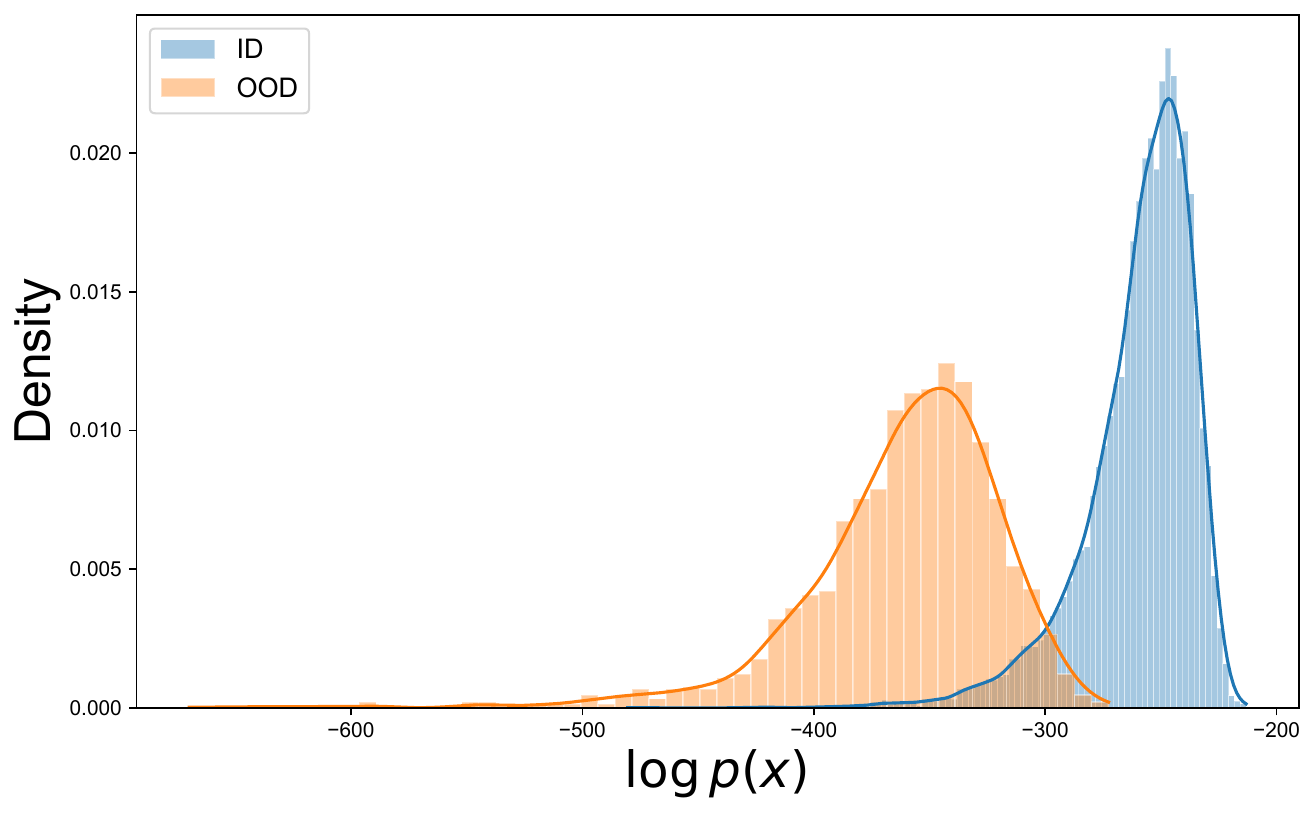}
            \caption*{}
            \label{fig:cifar10_7}
        \end{subfigure}
        \begin{subfigure}[t]{0.45\textwidth}
            \centering
            \includegraphics[width=\textwidth, height=2.5cm]{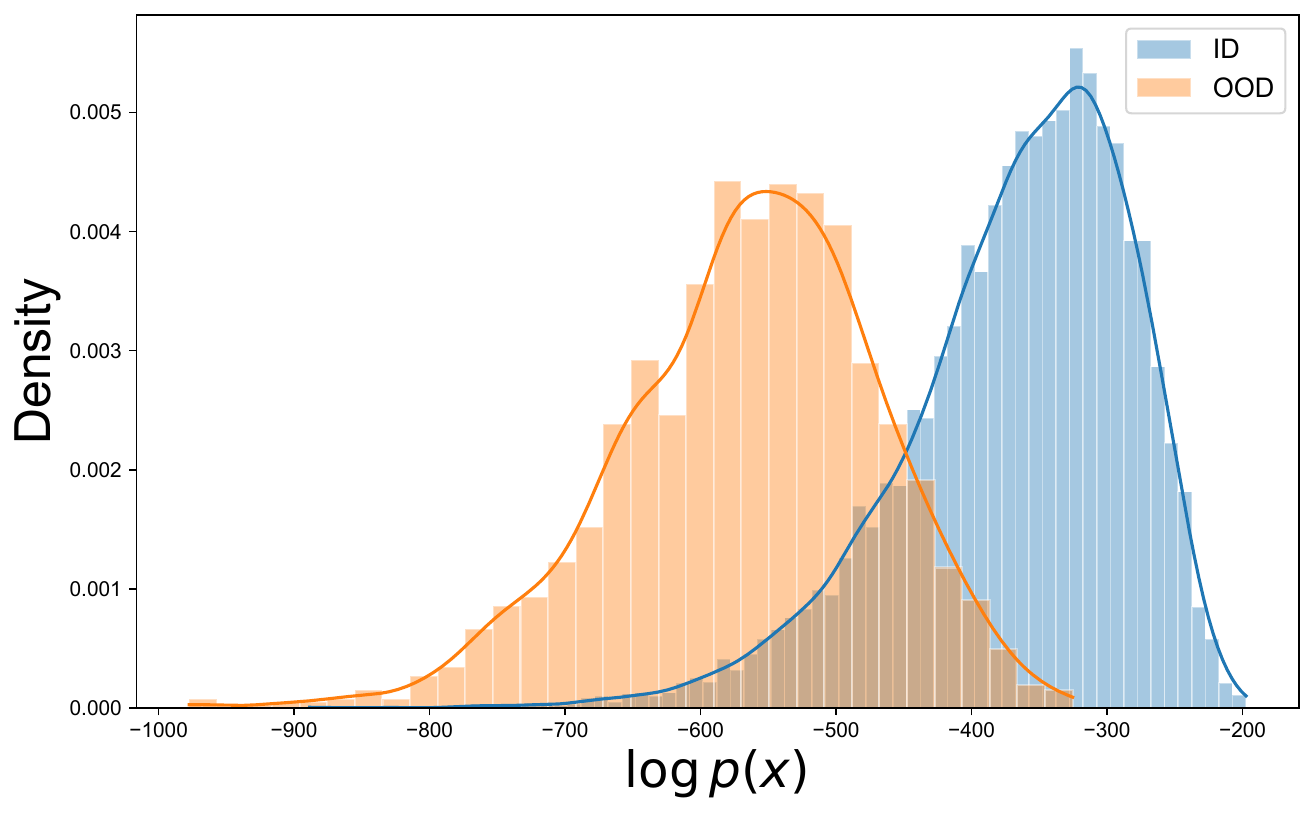}
            \caption*{ResNet18}
            \label{fig:cifar10_3_cifar100}
        \end{subfigure}%
        \begin{subfigure}[t]{0.45\textwidth}
            \centering
            \includegraphics[width=\textwidth, height=2.5cm]{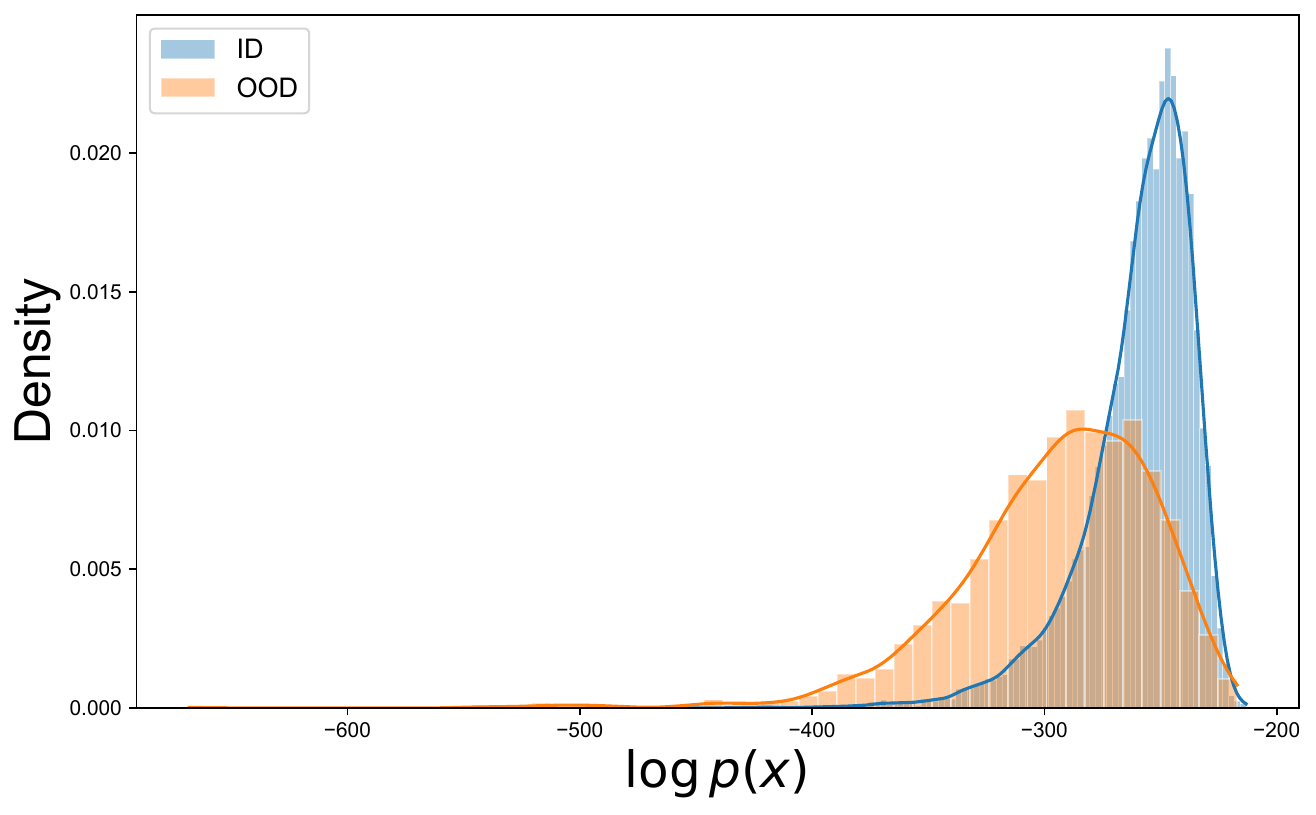}
            \caption*{WideResNet}
            \label{fig:cifar10_7_cifar100}
        \end{subfigure}
        \\
        \subcaption{$D_{in}=$ CIFAR-10. \textbf{Top} row represent Far-OOD where $D_{ood}$ is external OOD dataset. \textbf{Bottom} row indicates Mixed-OOD where $D_{ood}$ is CIFAR-100.}
        \label{fig:far}
    \end{minipage}
    \hfill
    \begin{minipage}{.45\textwidth}
        \centering
        \begin{subfigure}[t]{0.45\textwidth}
            \centering
            \includegraphics[width=\textwidth, height=2.5cm]{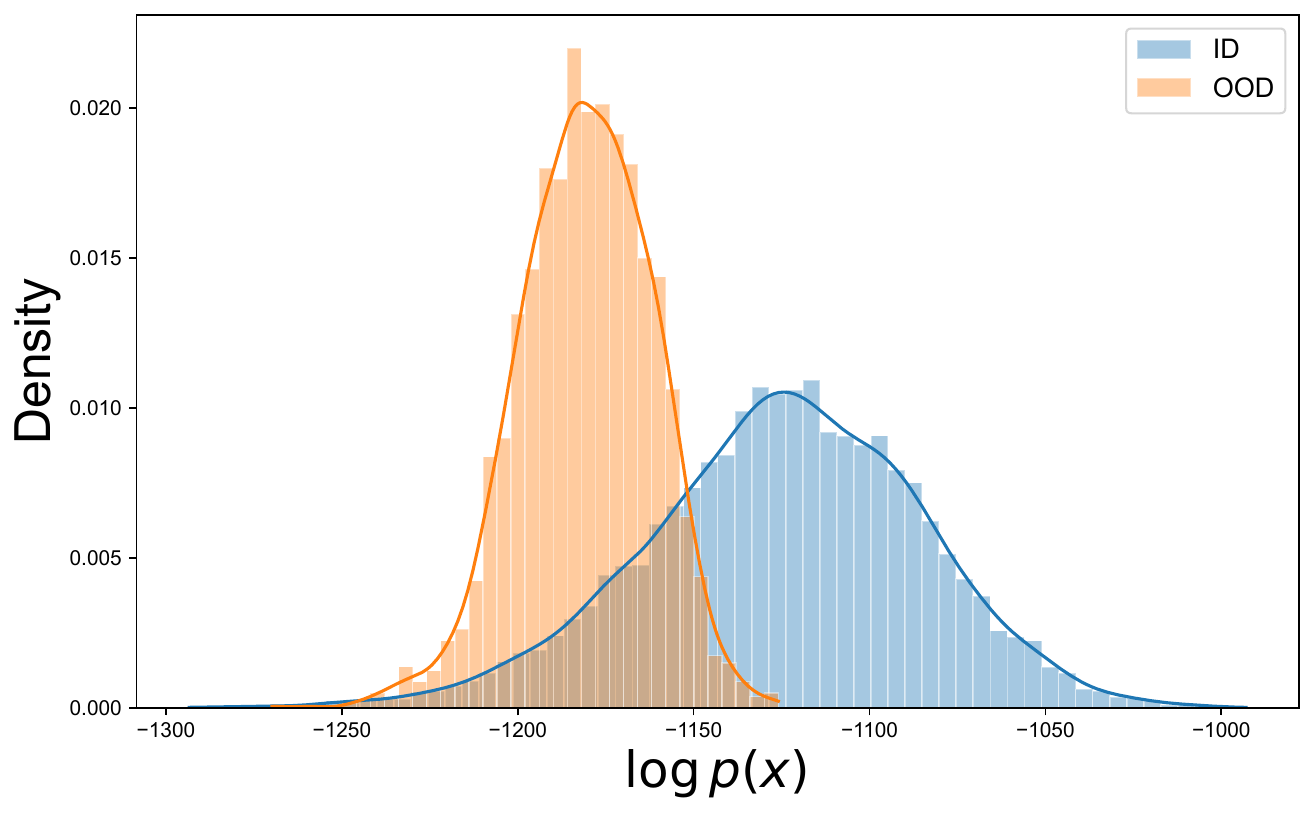}
            \caption*{}
            \label{fig:cifar100_5}
        \end{subfigure}%
        \begin{subfigure}[t]{0.45\textwidth}
            \centering
            \includegraphics[width=\textwidth, height=2.5cm]{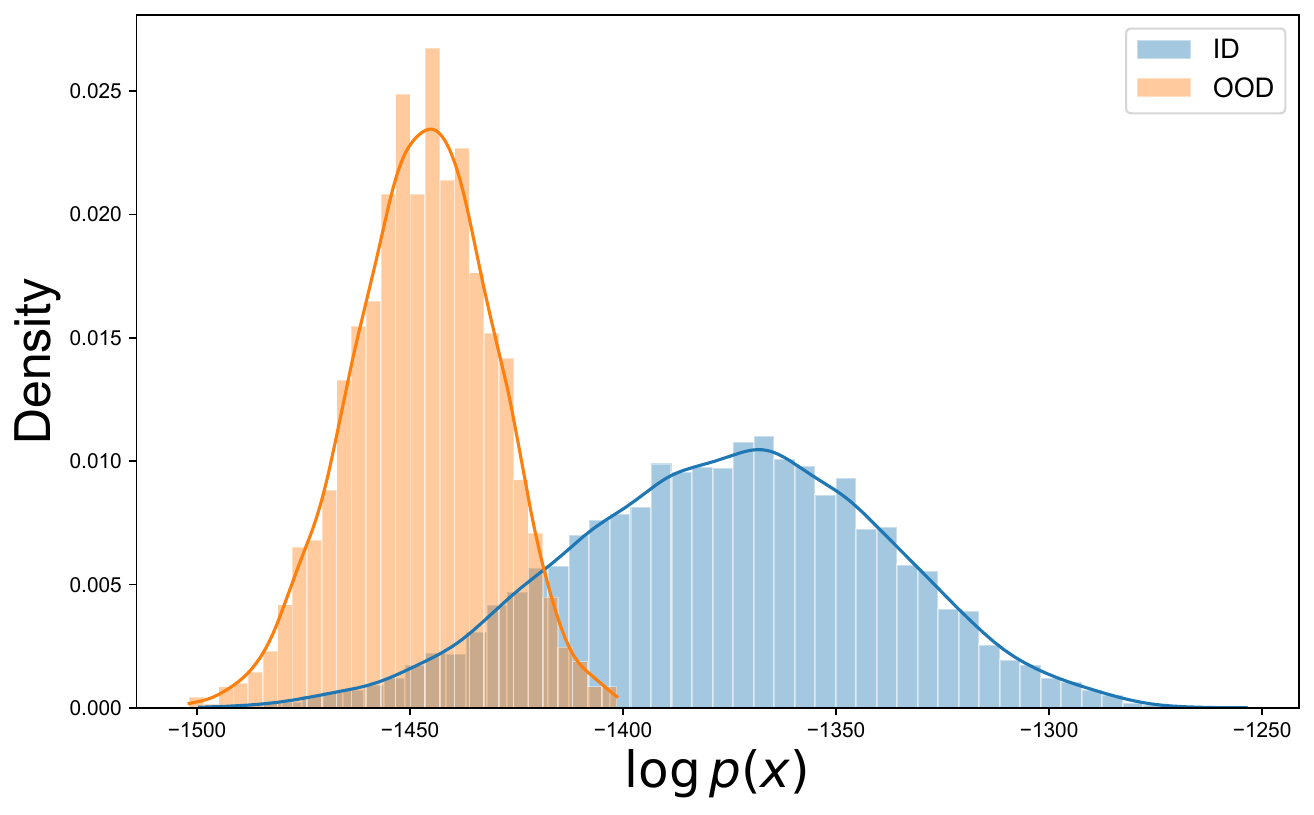}
            \caption*{}
            \label{fig:cifar100_6}
        \end{subfigure}
         \begin{subfigure}[t]{0.45\textwidth}
            \centering
            \includegraphics[width=\textwidth, height=2.5cm]{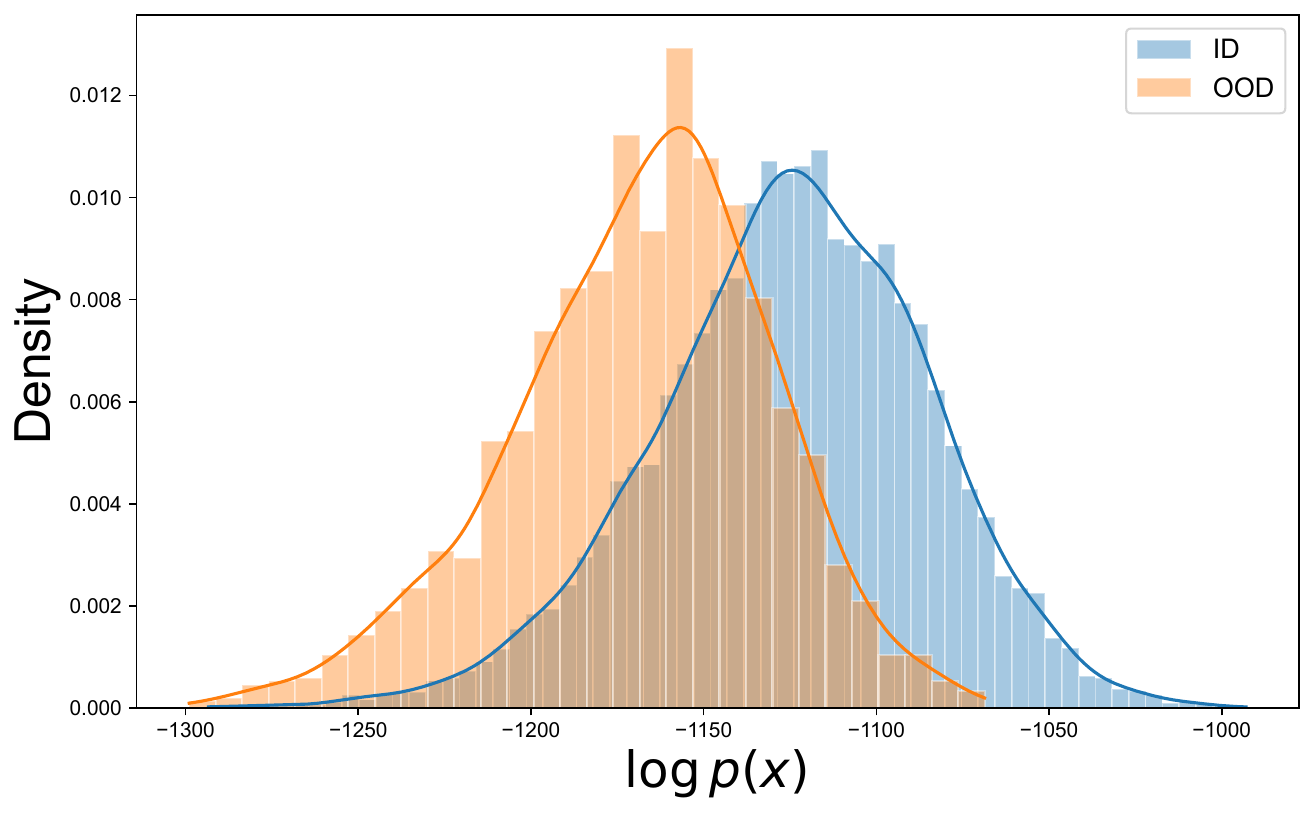}
            \caption*{ResNet18}
            \label{fig:cifar100_5_cifar10}
        \end{subfigure}%
        \begin{subfigure}[t]{0.45\textwidth}
            \centering
            \includegraphics[width=\textwidth, height=2.5cm]{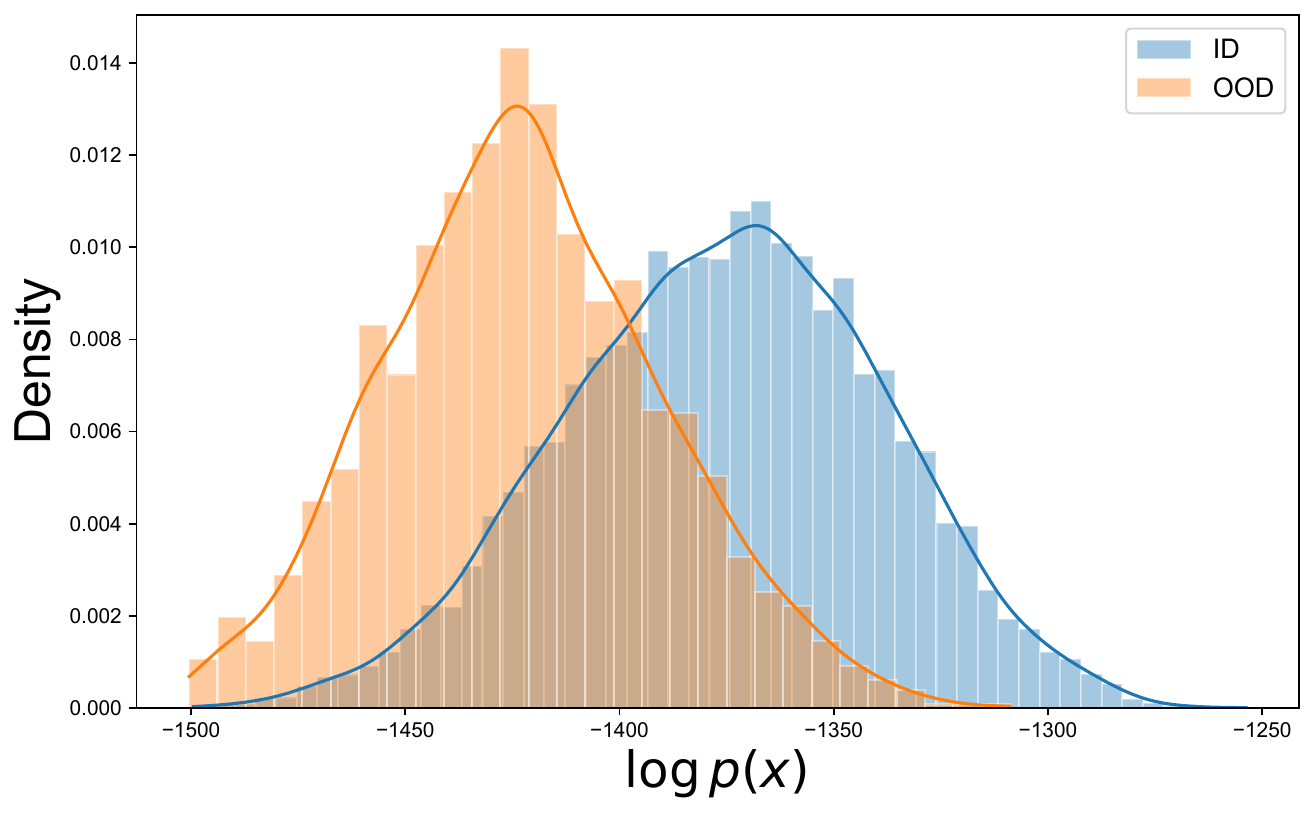}
            \caption*{WideResNet}
            \label{fig:cifar100_6_cifar10}
        \end{subfigure}\\
        \subcaption{$D_{in}=$ CIFAR-100. \textbf{Top} row represent Far-OOD where $D_{ood}$ is external OOD dataset. \textbf{Bottom} row indicates Near-OOD where $D_{ood}$ is CIFAR-10.}
        \label{fig:near}
    \end{minipage}
    \caption{Log-likelihood plots of trained \textit{FlowCon} }
    \label{fig:likelihood}
    \vspace{-5mm}
\end{figure}

As we move across different OOD spectrums (Tables \ref{tab:far_ood_results} - \ref{tab:near_ood_results}), we observe a gradual decrease in performance. Since \textit{FlowCon} is modelled on probability densities, it allows us to effectively understand and visualize the impact of the OOD spectrum by plotting the histogram of log-likelihood. Figs. \ref{fig:far} and \ref{fig:near} plot likelihood values for CIFAR-10 and CIFAR-100 respectively. For each figure, the top row presents the far-OOD context. For Fig. \ref{fig:far}, the bottom row presents the mixed-OOD context where \textit{FlowCon} trained on CIFAR-10 is evaluated on CIFAR-100 test data as OOD. Conversely, in Fig. \ref{fig:near} the bottom presents near-OOD context where \textit{FlowCon} is trained on CIFAR-100 and evaluated on CIFAR-10 as OOD.

For both Figs. \ref{fig:far} and \ref{fig:near}, as we move from top to bottom row, it is apparent that the overlap between likelihood plots increases. This is in agreement with the performance reported in Section \ref{subsec:res} as we observe a decline in metrics under more challenging OOD contexts. However, it is crucial to highlight that even under near-OOD conditions, the highest likelihood of falsely accepted OOD samples never exceeds that of the highest accepted ID sample. This aspect of \textit{FlowCon} is pivotal to its robust performance since it addresses an important issue of flow models described by Kirichenko \etal \cite{kirichenko2020normalizing} where normalizing flow models assign highest likelihood to OOD samples regardless of its training dataset.

\vspace{-4mm}
\subsection{Comparison with ResFlow}
\label{subsec:resflo}
\vspace{-2mm}

\begin{figure}
    \centering
    \begin{minipage}{0.48\textwidth}
        \centering
        \begin{subfigure}[t]{0.45\textwidth}
            \centering
            \includegraphics[width=\textwidth, height=2.5cm]{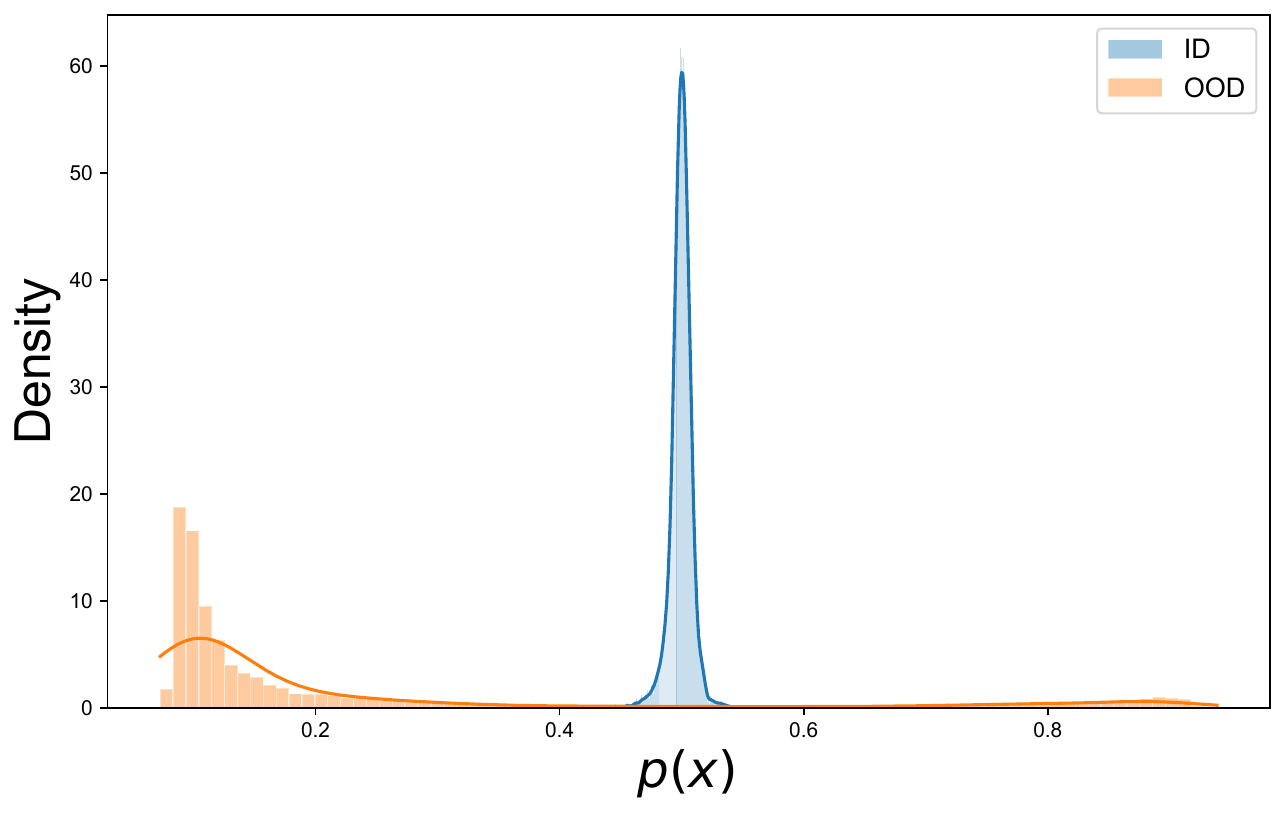}
            \caption{Far-OOD}
            \label{fig:cifar10_3}
        \end{subfigure}%
       \begin{subfigure}[t]{0.45\textwidth}
            \centering
            \includegraphics[width=\textwidth, height=2.5cm]{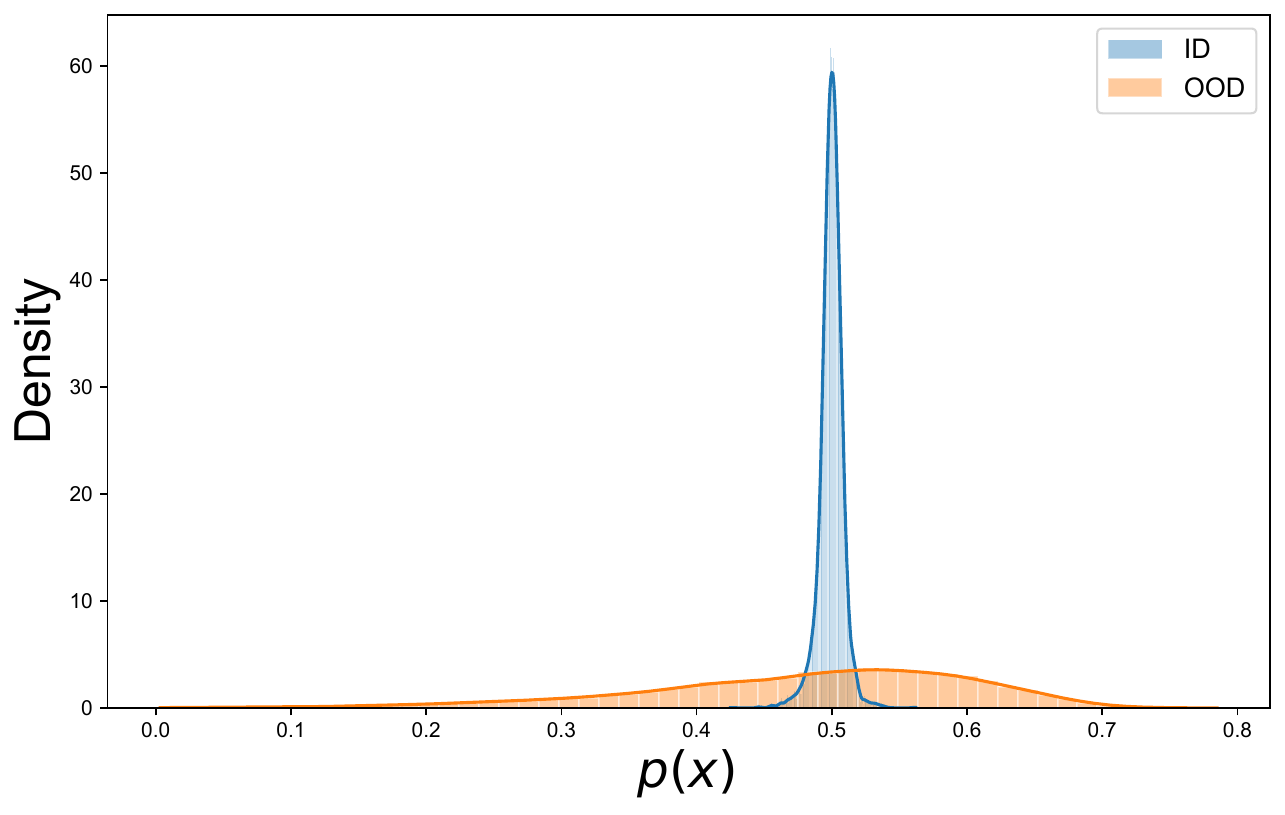}
            \caption{Near-OOD}
            \label{fig:cifar10_3_cifar100}
        \end{subfigure}
        \\
        \subcaption*{ResNet18}
    \end{minipage}
    \hfill
    \begin{minipage}{.48\textwidth}
        \centering
         \begin{subfigure}[t]{0.45\textwidth}
            \centering
            \includegraphics[width=\textwidth, height=2.5cm]{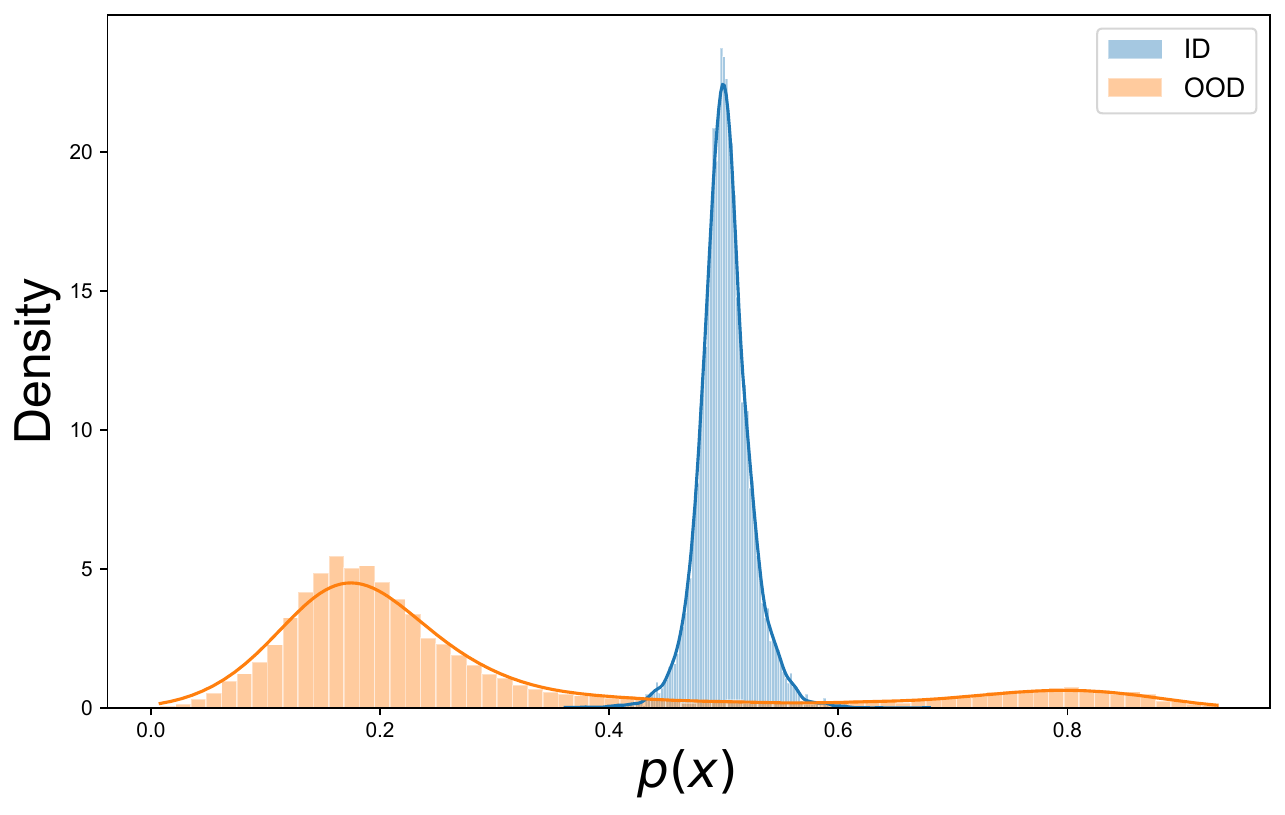}
            \caption{Far-OOD}
            \label{fig:cifar10_7}
        \end{subfigure}
        \begin{subfigure}[t]{0.45\textwidth}
            \centering
            \includegraphics[width=\textwidth, height=2.5cm]{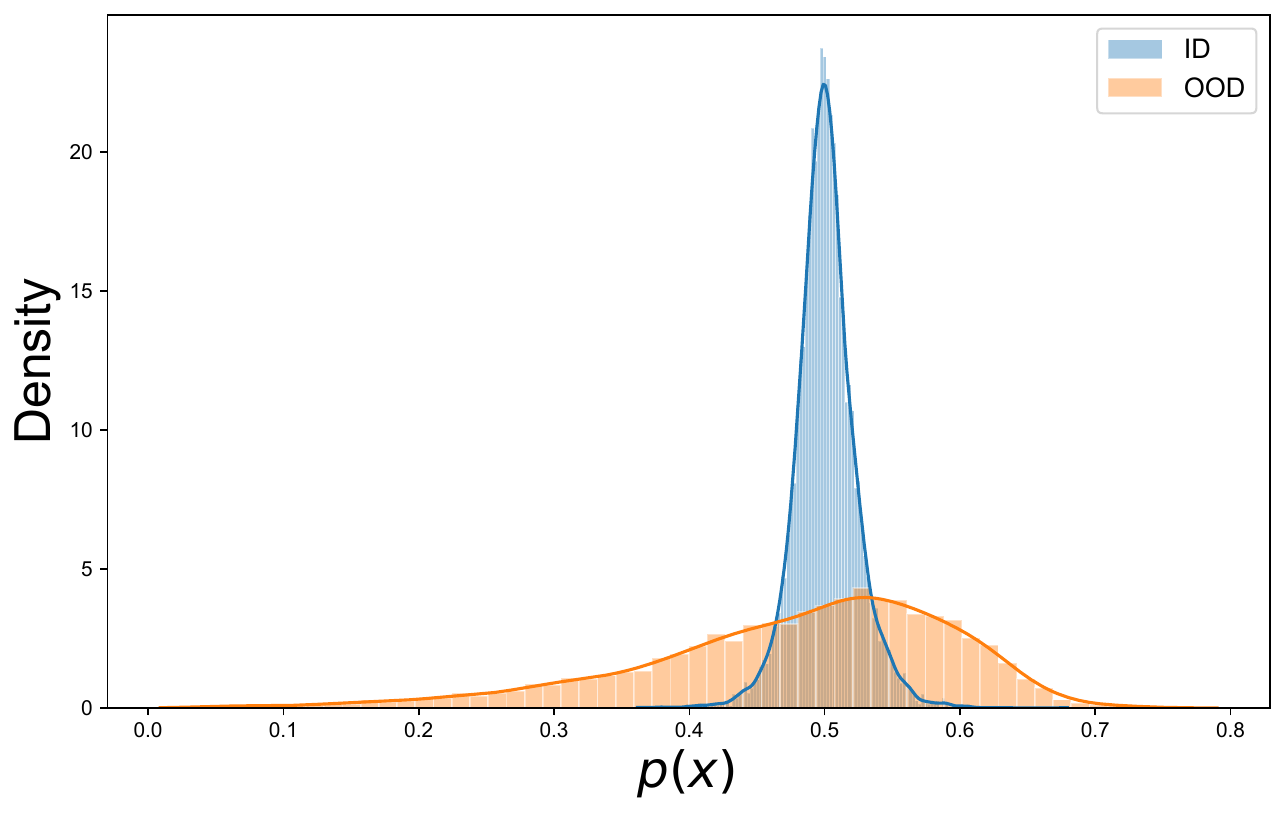}
            \caption{Near-OOD}
            \label{fig:cifar10_7_cifar100}
        \end{subfigure}
        \\
        \subcaption*{WideResNet}
    \end{minipage}
    \caption{Histogram plots on regressor scores of CIFAR-100 trained on ResFlow.}
    \label{fig:resflow_ll}
    \vspace{-3mm}
\end{figure}

\textit{FlowCon} addresses a critical constraint of ResFlow \cite{zisselman2020deep} models. Unlike ResFlows, our training framework is independent of the number of layers of the classifier or the number of classes in the ID dataset. Since ResFlow models all the intermediate features of the classifier, we evaluate the regressor scores assigned on near-OOD context. 
Fig. \ref{fig:resflow_ll} plots the histogram of the scores predicted by the ResFlow model on CIFAR-100 as ID and CIFAR-10 as OOD dataset. Due to a well calibrated regressor, the scores for the ID dataset have a low variance. However, unlike the likelihood plots for \textit{FlowCon}, ResFlows assigns the highest likelihood for OOD samples and not the ID samples. 

\subsection{\textit{FlowCon} as Classifier}
Since we train \textit{FlowCon} in a contrastive manner, we hypothesize that the emperical class-wise distributions $\mathcal{N}_{\mathcal{X}}$ computed in Section \ref{subsec:inference} captures the discriminative information of the original classifier. To test this, we predict the class of a given test sample, $z_{test}$ using Bayes' decision rule as  
\begin{equation}
    y_{test} = \arg \max_{i\in \{1, ..., k\}}p_Z(z_{test}|\mathcal{N}_{y=i})
\end{equation}
Using this, we compute the image classification accuracy and compare it with the original pretrained classifier on CIFAR-10 and CIFAR-100 using both ResNet18 and WideResNet. Table \ref{tab:class} reports the accuracy scores. We observe that difference between \textit{FlowCon} and the original classifier is negligible, therefore the assertion that \textit{FlowCon} is a class-preserving approach remains valid. Furthermore, in the case of WideResNet, our approach marginally outperforms the original classifier. This implies that a single branch suffices in both OOD detection and ID classification.
\begin{table}
\vspace{-5mm}
    \begin{center}
    \caption{Class-preserving property of \textit{FlowCon}. The classification accuracy of our approach remains closely bounded to the original classifier.}
    \label{tab:class}
    \small{
    \begin{adjustbox}{width=0.8\textwidth, totalheight=0.2\textheight, keepaspectratio}
        \begin{tabular}{cclc}
        \toprule
            \multirow{2}{*}{$D_{in}$} & \multirow{2}{*}{Model} & \multirow{2}{*}{Method} & \multirow{2}{*}{Accuracy $\uparrow$}  \\
             & & & \\
            \midrule
            \multirow{2}{*}{\makecell{CIFAR-10}} & \multirow{2}{*}{ResNet18} & Orig  & 94.3  \\
            & & \textbf{FlowCon}  & 94.2 \\
            \midrule 
            \multirow{2}{*}{\makecell{CIFAR-10 }} & \multirow{2}{*}{WideResNet} & Orig  & 93.3  \\
            & & \textbf{FlowCon } & 93.8 \\
            \midrule 
            \multirow{2}{*}{\makecell{CIFAR-100 }} & \multirow{2}{*}{ResNet18} & Orig  & 75.8  \\
            & & \textbf{FlowCon}  & 74.9 \\
            \midrule 
            \multirow{2}{*}{\makecell{CIFAR-100 }} & \multirow{2}{*}{WideResNet} & Orig  & 70.9  \\
            & & \textbf{FlowCon}  & 71.1 \\
    
        \bottomrule
        \end{tabular}

    \end{adjustbox}
    }
    \end{center}
    \vspace{-14mm}
\end{table}

\section{Discussion}
\label{sec:disc}

\subsubsection{Latent Space Visualization.}
   
\begin{figure}
\vspace{-3mm}
    \centering
    \begin{minipage}{0.48\textwidth}
        \centering
        \begin{subfigure}[t]{0.48\textwidth}
            \centering
            \includegraphics[width=0.95\textwidth, height=2.5cm]{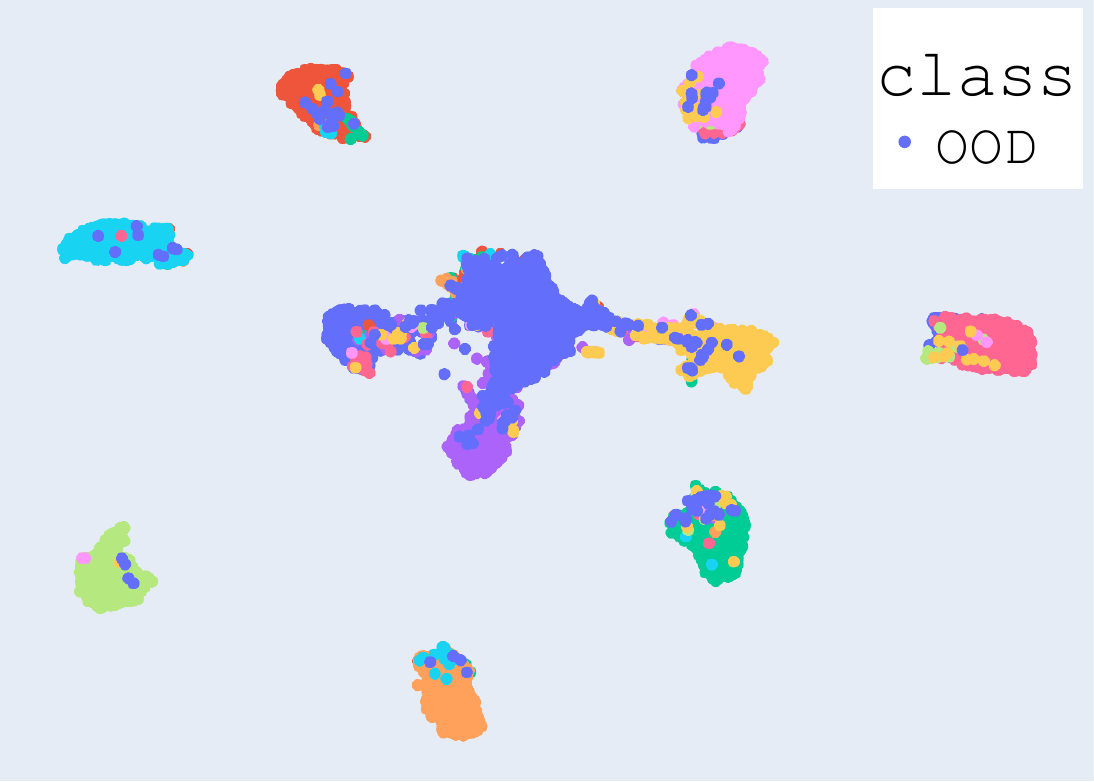}
            \caption{Far-OOD}
        \end{subfigure}%
        \begin{subfigure}[t]{0.48\textwidth}
            \centering
            \includegraphics[width=0.95\textwidth, height=2.5cm]{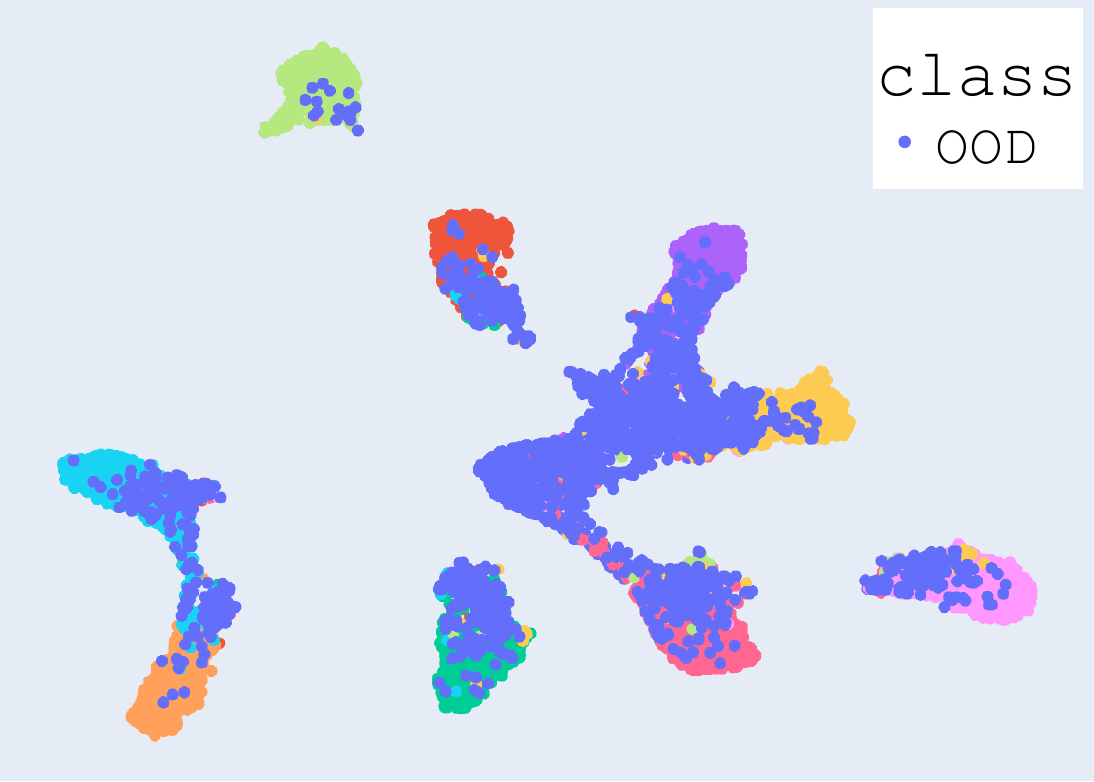}
            \caption{Near-OOD}
        \end{subfigure}
        \\
        \subcaption*{ResNet18}
    \end{minipage}
    \hfill
    \begin{minipage}{.48\textwidth}
        \centering
        \begin{subfigure}[t]{0.48\textwidth}
            \centering
            \includegraphics[width=0.95\textwidth, height=2.5cm]{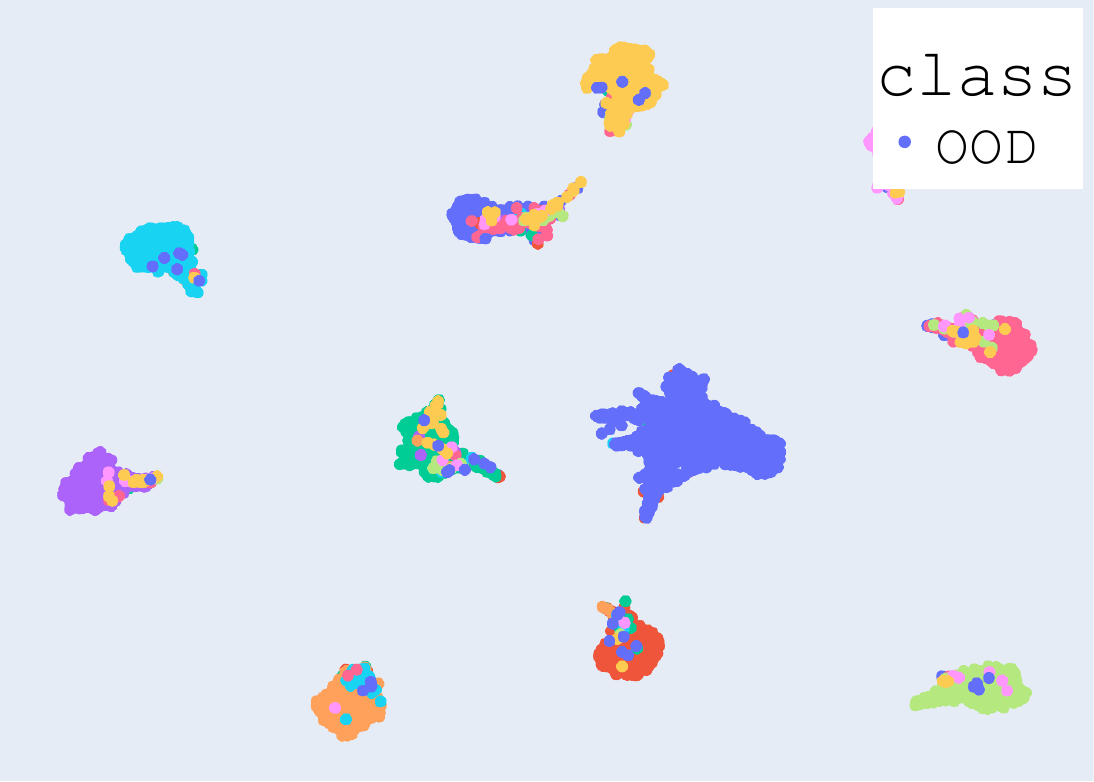}
            \caption{Far-OOD}
            \label{fig:umap_wrn_far}
        \end{subfigure}%
        \begin{subfigure}[t]{0.48\textwidth}
            \centering
            \includegraphics[width=0.95\textwidth, height=2.5cm]{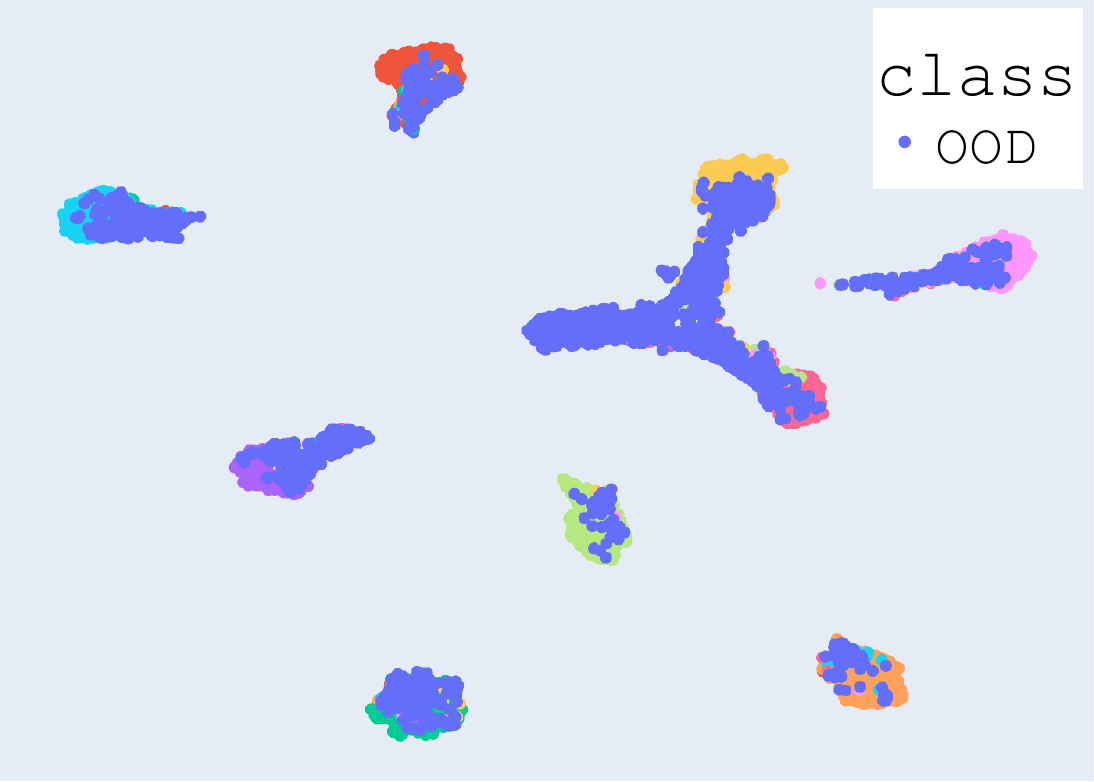}
            \caption{Near-OOD}
            \label{fig:umap_wrn_near}
        \end{subfigure}
        \\
        \subcaption*{WideResNet}
    \end{minipage}
    \caption{UMAP embeddings of $z_{flow}$ trained on CIFAR-10 using \textit{FlowCon} }
    \label{fig:umap}
    \vspace{-10mm}
\end{figure}

Apart from visualizing the density plots, we plot the low-dimensional UMAP embeddings \cite{mcinnes2018umap} of the learned features $z_{flow}$. We show the plots for CIFAR-10 trained on both ResNet18 and WideResNet in Fig. \ref{fig:umap}. The blue color represents the OOD data. For far-OOD, we use SVHN \cite{netzer2011reading} and for near-OOD, we use CIFAR-100. The UMAP embeddings exhibit a well-clustered latent space that further supports the classification ability of \textit{FlowCon}. Moreover, for near-OOD contexts, we can observe the ID class clusters overlap with OOD data with similar semantics. This is analogous to overlapping of the likelihood plots as shown in Fig. \ref{fig:likelihood} (bottom rows). 

\subsubsection{Impact of $\lambda$}
Nalisnick \etal \cite{nalisnick2019hybrid} first explored the role of $\lambda$ in the context of flow-based classification. The value $\lambda=1/d$, where $d$ is the dimension of $z_{flow}$ was found to be most suitable for OOD detection. In case of $FlowCon$, we found $0.07$ to be the most suitable value optimizing both $\mathcal{L}_{flow}$  and $\mathcal{L}_{con}$. We experiment with different $\lambda$ values in the range of $[0.05, 1]$ for WideResNet trained on CIFAR-100 and report the results in Table \ref{tab:lambda}. The table demonstrates that an increasing $\lambda$ value reduces the overall performance of \textit{FlowCon}.
\begin{table}
\vspace{-3mm}
    \setlength\tabcolsep{1pt}
    \begin{center}
    \caption{Effect of $\lambda$ in optimizing $\mathcal{L}_{flow}$ and $ \mathcal{L}_{con}$ under far-OOD context. Shaded region reports $\lambda$ values used in experiments.}
    \label{tab:lambda}
    \vskip -1em
    \begin{adjustbox}{width=0.5\textwidth, totalheight=0.4\textheight, keepaspectratio}
    \begin{tabular}{clcccc}
    \toprule
        $D_{in}$ & \multirow{2}{*}{$\lambda$} & \multirow{2}{*}{AUROC $\uparrow$} & \multirow{2}{*}{AUPR-S $\uparrow$} & \multirow{2}{*}{AUPR-E $\uparrow$} & \multirow{2}{*}{FPR-95 $\downarrow$} \\
        (model) & & & & & \\
        \midrule
        \multirow{5}{*}{\makecell{CIFAR-100 \\ (WideResNet)}} & $0.05$  & 75.62 & 92.7 & 41.84 & 72.58 \\
        
        & \cellcolor{LightGray}$0.07$  & \cellcolor{LightGray}83.62 & \cellcolor{LightGray}96.60 & \cellcolor{LightGray}53.34 & \cellcolor{LightGray}60.28 \\
        & $0.3$  & 75.75 & 92.76 & 48.61 & 63.67 \\
        & $0.5$  & 78.60 & 93.96 & 49.07 & 65.92 \\
        & $1.0$  & 78.57 & 93.24 & 45.94 & 67.85 \\
        
    \bottomrule
    \end{tabular}
    \end{adjustbox}
    \end{center}
    \vspace{-11mm}
\end{table}

\subsubsection{Limitations and Future Work.} 
One of the constraints of normalizing flows \cite{papamakarios2021normalizing} is that the dimensions of input ($z_{emb}$) and output ($z_{flow}$) should be the same. This potentially enforces the model to operate on low dimensional feature vectors depending on the classifier network being used, as observed in experiments pertaining to WideResNet. We emphasize that the reduced dimensionality constrains the learning of \textit{FlowCon}, which we will address in future work.

\section{Conclusion}
\vspace{-3mm}
A new approach to OOD detection called \textit{FlowCon} was proposed, which does not use an external dataset as OOD or retrain the original classifier. The key intuition of our approach is that the class-informed density estimator can recognize OOD data simply by filtering out low density samples. The proposed approach operates on deep features, instead of the raw input space, and therefore can be extended to different domains. The best results were obtained on ResNet18 features on all OOD contexts and exhibited competitive performance on WideResNet features.


%
%
\bibliographystyle{splncs04}
\bibliography{egbib}

\begin{thebibliography}{10}
\providecommand{\url}[1]{\texttt{#1}}
\providecommand{\urlprefix}{URL }
\providecommand{\doi}[1]{https://doi.org/#1}

\bibitem{abati2019latent}
Abati, D., Porrello, A., Calderara, S., Cucchiara, R.: Latent space autoregression for novelty detection. In: Proceedings of the IEEE/CVF conference on computer vision and pattern recognition. pp. 481--490 (2019)

\bibitem{amodei2016concrete}
Amodei, D., Olah, C., Steinhardt, J., Christiano, P., Schulman, J., Man{\'e}, D.: Concrete problems in ai safety. arXiv preprint arXiv:1606.06565  (2016)

\bibitem{bhattacharyya1946measure}
Bhattacharyya, A.: On a measure of divergence between two multinomial populations. Sankhy{\=a}: the indian journal of statistics  (1946)

\bibitem{chen2019residual}
Chen, R.T., Behrmann, J., Duvenaud, D.K., Jacobsen, J.H.: Residual flows for invertible generative modeling. Advances in Neural Information Processing Systems  \textbf{32} (2019)

\bibitem{chen2020simple}
Chen, T., Kornblith, S., Norouzi, M., Hinton, G.: A simple framework for contrastive learning of visual representations. In: International conference on machine learning. pp. 1597--1607. PMLR (2020)

\bibitem{cimpoi2014describing}
Cimpoi, M., Maji, S., Kokkinos, I., Mohamed, S., Vedaldi, A.: Describing textures in the wild. In: Proceedings of the IEEE conference on computer vision and pattern recognition. pp. 3606--3613 (2014)

\bibitem{deecke2019image}
Deecke, L., Vandermeulen, R., Ruff, L., Mandt, S., Kloft, M.: Image anomaly detection with generative adversarial networks. In: Machine Learning and Knowledge Discovery in Databases: European Conference, ECML PKDD 2018, Dublin, Ireland, September 10--14, 2018, Proceedings, Part I 18. pp. 3--17. Springer (2019)

\bibitem{dinh2016density}
Dinh, L., Sohl-Dickstein, J., Bengio, S.: Density estimation using real nvp. arXiv preprint arXiv:1605.08803  (2016)

\bibitem{filos2020can}
Filos, A., Tigkas, P., McAllister, R., Rhinehart, N., Levine, S., Gal, Y.: Can autonomous vehicles identify, recover from, and adapt to distribution shifts? In: International Conference on Machine Learning. pp. 3145--3153. PMLR (2020)

\bibitem{frosst2019analyzing}
Frosst, N., Papernot, N., Hinton, G.: Analyzing and improving representations with the soft nearest neighbor loss. In: International conference on machine learning. pp. 2012--2020. PMLR (2019)

\bibitem{goodfellow2014explaining}
Goodfellow, I.J., Shlens, J., Szegedy, C.: Explaining and harnessing adversarial examples. arXiv preprint arXiv:1412.6572  (2014)

\bibitem{he2016deep}
He, K., Zhang, X., Ren, S., Sun, J.: Deep residual learning for image recognition. In: Proceedings of the IEEE conference on computer vision and pattern recognition. pp. 770--778 (2016)

\bibitem{hendrycks2021unsolved}
Hendrycks, D., Carlini, N., Schulman, J., Steinhardt, J.: Unsolved problems in ml safety. arXiv preprint arXiv:2109.13916  (2021)

\bibitem{hendrycks2016baseline}
Hendrycks, D., Gimpel, K.: A baseline for detecting misclassified and out-of-distribution examples in neural networks. arXiv preprint arXiv:1610.02136  (2016)

\bibitem{hendrycks2022x}
Hendrycks, D., Mazeika, M.: X-risk analysis for ai research. arXiv preprint arXiv:2206.05862  (2022)

\bibitem{hendrycks2018deep}
Hendrycks, D., Mazeika, M., Dietterich, T.: Deep anomaly detection with outlier exposure. arXiv preprint arXiv:1812.04606  (2018)

\bibitem{hornauer2023heatmap}
Hornauer, J., Belagiannis, V.: Heatmap-based out-of-distribution detection. In: Proceedings of the IEEE/CVF Winter Conference on Applications of Computer Vision. pp. 2603--2612 (2023)

\bibitem{khosla2020supervised}
Khosla, P., Teterwak, P., Wang, C., Sarna, A., Tian, Y., Isola, P., Maschinot, A., Liu, C., Krishnan, D.: Supervised contrastive learning. Advances in neural information processing systems  \textbf{33},  18661--18673 (2020)

\bibitem{kingma2014adam}
Kingma, D.P., Ba, J.: Adam: A method for stochastic optimization. arXiv preprint arXiv:1412.6980  (2014)

\bibitem{kingma2018glow}
Kingma, D.P., Dhariwal, P.: Glow: Generative flow with invertible 1x1 convolutions. Advances in neural information processing systems  \textbf{31} (2018)

\bibitem{kingma2014semi}
Kingma, D.P., Mohamed, S., Jimenez~Rezende, D., Welling, M.: Semi-supervised learning with deep generative models. Advances in neural information processing systems  \textbf{27} (2014)

\bibitem{kirichenko2020normalizing}
Kirichenko, P., Izmailov, P., Wilson, A.G.: Why normalizing flows fail to detect out-of-distribution data. Advances in neural information processing systems  \textbf{33},  20578--20589 (2020)

\bibitem{krizhevsky2009learning}
Krizhevsky, A., Hinton, G., et~al.: Learning multiple layers of features from tiny images  (2009)

\bibitem{lee2018simple}
Lee, K., Lee, K., Lee, H., Shin, J.: A simple unified framework for detecting out-of-distribution samples and adversarial attacks. Advances in neural information processing systems  \textbf{31} (2018)

\bibitem{liang2017enhancing}
Liang, S., Li, Y., Srikant, R.: Enhancing the reliability of out-of-distribution image detection in neural networks. arXiv preprint arXiv:1706.02690  (2017)

\bibitem{liu2020energy}
Liu, W., Wang, X., Owens, J., Li, Y.: Energy-based out-of-distribution detection. Advances in neural information processing systems  \textbf{33},  21464--21475 (2020)

\bibitem{lu2023uncertainty}
Lu, F., Zhu, K., Zhai, W., Zheng, K., Cao, Y.: Uncertainty-aware optimal transport for semantically coherent out-of-distribution detection. In: Proceedings of the IEEE/CVF Conference on Computer Vision and Pattern Recognition. pp. 3282--3291 (2023)

\bibitem{mcinnes2018umap}
McInnes, L., Healy, J., Melville, J.: Umap: Uniform manifold approximation and projection for dimension reduction. arXiv preprint arXiv:1802.03426  (2018)

\bibitem{mohseni2020self}
Mohseni, S., Pitale, M., Yadawa, J., Wang, Z.: Self-supervised learning for generalizable out-of-distribution detection. In: Proceedings of the AAAI Conference on Artificial Intelligence. vol.~34, pp. 5216--5223 (2020)

\bibitem{morteza2022provable}
Morteza, P., Li, Y.: Provable guarantees for understanding out-of-distribution detection. In: Proceedings of the AAAI Conference on Artificial Intelligence. vol.~36, pp. 7831--7840 (2022)

\bibitem{nalisnick2019hybrid}
Nalisnick, E., Matsukawa, A., Teh, Y.W., Gorur, D., Lakshminarayanan, B.: Hybrid models with deep and invertible features. In: International Conference on Machine Learning. pp. 4723--4732. PMLR (2019)

\bibitem{netzer2011reading}
Netzer, Y., Wang, T., Coates, A., Bissacco, A., Wu, B., Ng, A.Y.: Reading digits in natural images with unsupervised feature learning  (2011)

\bibitem{papamakarios2021normalizing}
Papamakarios, G., Nalisnick, E., Rezende, D.J., Mohamed, S., Lakshminarayanan, B.: Normalizing flows for probabilistic modeling and inference. The Journal of Machine Learning Research  \textbf{22}(1),  2617--2680 (2021)

\bibitem{pidhorskyi2018generative}
Pidhorskyi, S., Almohsen, R., Doretto, G.: Generative probabilistic novelty detection with adversarial autoencoders. Advances in neural information processing systems  \textbf{31} (2018)

\bibitem{reyes2023testing}
Reyes-Gonz{\'a}lez, H., Torre, R.: Testing the boundaries: Normalizing flows for higher dimensional data sets. In: Journal of Physics: Conference Series. vol.~2438, p. 012155. IOP Publishing (2023)

\bibitem{roy2022does}
Roy, A.G., Ren, J., Azizi, S., Loh, A., Natarajan, V., Mustafa, B., Pawlowski, N., Freyberg, J., Liu, Y., Beaver, Z., et~al.: Does your dermatology classifier know what it doesn’t know? detecting the long-tail of unseen conditions. Medical Image Analysis  \textbf{75},  102274 (2022)

\bibitem{sabokrou2018adversarially}
Sabokrou, M., Khalooei, M., Fathy, M., Adeli, E.: Adversarially learned one-class classifier for novelty detection. In: Proceedings of the IEEE conference on computer vision and pattern recognition. pp. 3379--3388 (2018)

\bibitem{shafaei2018less}
Shafaei, A., Schmidt, M., Little, J.J.: A less biased evaluation of out-of-distribution sample detectors. arXiv preprint arXiv:1809.04729  (2018)

\bibitem{sun2021react}
Sun, Y., Guo, C., Li, Y.: React: Out-of-distribution detection with rectified activations. Advances in Neural Information Processing Systems  \textbf{34},  144--157 (2021)

\bibitem{tabak2013family}
Tabak, E.G., Turner, C.V.: A family of nonparametric density estimation algorithms. Communications on Pure and Applied Mathematics  \textbf{66}(2),  145--164 (2013)

\bibitem{torralba200880}
Torralba, A., Fergus, R., Freeman, W.T.: 80 million tiny images: A large data set for nonparametric object and scene recognition. IEEE transactions on pattern analysis and machine intelligence  \textbf{30}(11),  1958--1970 (2008)

\bibitem{wang2017chestx}
Wang, X., Peng, Y., Lu, L., Lu, Z., Bagheri, M., Summers, R.M.: Chestx-ray8: Hospital-scale chest x-ray database and benchmarks on weakly-supervised classification and localization of common thorax diseases. In: Proceedings of the IEEE conference on computer vision and pattern recognition. pp. 2097--2106 (2017)

\bibitem{wen2022self}
Wen, X., Zhao, B., Zheng, A., Zhang, X., Qi, X.: Self-supervised visual representation learning with semantic grouping. Advances in Neural Information Processing Systems  \textbf{35},  16423--16438 (2022)

\bibitem{winkens2020contrastive}
Winkens, J., Bunel, R., Roy, A.G., Stanforth, R., Natarajan, V., Ledsam, J.R., MacWilliams, P., Kohli, P., Karthikesalingam, A., Kohl, S., et~al.: Contrastive training for improved out-of-distribution detection. arXiv preprint arXiv:2007.05566  (2020)

\bibitem{xu2015turkergaze}
Xu, P., Ehinger, K.A., Zhang, Y., Finkelstein, A., Kulkarni, S.R., Xiao, J.: Turkergaze: Crowdsourcing saliency with webcam based eye tracking. arXiv preprint arXiv:1504.06755  (2015)

\bibitem{yang2021semantically}
Yang, J., Wang, H., Feng, L., Yan, X., Zheng, H., Zhang, W., Liu, Z.: Semantically coherent out-of-distribution detection. In: Proceedings of the IEEE/CVF International Conference on Computer Vision. pp. 8301--8309 (2021)

\bibitem{yu2015lsun}
Yu, F., Seff, A., Zhang, Y., Song, S., Funkhouser, T., Xiao, J.: Lsun: Construction of a large-scale image dataset using deep learning with humans in the loop. arXiv preprint arXiv:1506.03365  (2015)

\bibitem{zagoruyko2016wide}
Zagoruyko, S., Komodakis, N.: Wide residual networks. arXiv preprint arXiv:1605.07146  (2016)

\bibitem{zhang2020hybrid}
Zhang, H., Li, A., Guo, J., Guo, Y.: Hybrid models for open set recognition. In: Computer Vision--ECCV 2020: 16th European Conference, Glasgow, UK, August 23--28, 2020, Proceedings, Part III 16. pp. 102--117. Springer (2020)

\bibitem{zhou2017places}
Zhou, B., Lapedriza, A., Khosla, A., Oliva, A., Torralba, A.: Places: A 10 million image database for scene recognition. IEEE transactions on pattern analysis and machine intelligence  \textbf{40}(6),  1452--1464 (2017)

\bibitem{zisselman2020deep}
Zisselman, E., Tamar, A.: Deep residual flow for out of distribution detection. In: Proceedings of the IEEE/CVF Conference on Computer Vision and Pattern Recognition. pp. 13994--14003 (2020)

\bibitem{zong2018deep}
Zong, B., Song, Q., Min, M.R., Cheng, W., Lumezanu, C., Cho, D., Chen, H.: Deep autoencoding gaussian mixture model for unsupervised anomaly detection. In: International conference on learning representations (2018)

\end{thebibliography}
\end{document}